\documentclass{article}

\PassOptionsToPackage{numbers, compress}{natbib}
\usepackage[preprint]{neurips_2026}
\usepackage[most]{tcolorbox}
\usepackage[utf8]{inputenc}
\usepackage[T1]{fontenc}
\usepackage{hyperref}
\usepackage{url}
\usepackage{booktabs}
\usepackage{amsfonts}
\usepackage{amsmath}
\usepackage{amssymb}
\usepackage{nicefrac}
\usepackage{microtype}
\usepackage{xcolor}
\usepackage{algorithm}
\usepackage{algorithmic}
\usepackage{graphicx}
\usepackage{multirow}
\usepackage{enumitem}
\usepackage{mathtools}
\usepackage{float}
\usepackage{tikz}
\usepackage{tabularx}
\usepackage{makecell}
\usepackage{array}

\usetikzlibrary{positioning, arrows.meta, calc, fit, backgrounds, decorations.pathreplacing, shapes.geometric, shadows}
\usepackage{xcolor}

\definecolor{todocolor}{RGB}{255,200,0}

\newcommand{\method}{\textsc{$\mu$CRASP}}

\title{\textsc{$\mu$CRASP}: \underline{Mu}ltimodal \underline{C}hain-of-thought \underline{R}easoning \underline{a}ware \underline{S}tructured \underline{P}runing}

\author{%
Aritra Dutta \quad Somak Aditya \\
Indian Institute of Technology, Kharagpur\\
\texttt{ aritradutta24.23@kgpian.iitkgp.ac.in,saditya@cse.iitkgp.ac.in}
}

\begin{document}

\maketitle

\begin{abstract}
Vision-language models (VLMs) increasingly rely on chain-of-thought (CoT) reasoning to solve complex multimodal tasks, but their large parameter sizes make deployment comparably expensive. Structured model pruning methods offer a natural solution. However, existing methods fail to preserve CoT reasoning accuracy when applied to VLMs. We identify two reasons: i) CoT reasoning consistency is governed by sparse transition points (pivot tokens) in the generation trajectory, and conventional pruning methods are CoT-agnostic; and ii) traditional pruning designed for unimodal LLM's does not account for the activation distribution difference across the visual and textual modalities, making pruning significantly more challenging than in purely textual unimodal models. Based on this insight, we propose \method{}, a structured pruning framework that explicitly targets reasoning-critical components while preserving cross-modal alignment and accounting for layer-wise sensitivity under a global parameter budget. Our approach enables effective compression while maintaining both reasoning coherence and multimodal grounding. 
Experiments on four VLMs across three reasoning benchmarks show that \method{} consistently preserves reasoning quality under increasing compression. At $30\%$ pruning on Qwen2.5-VL-7B, CoT generated under \method{} retains a high consistency with original (dense) finetuned model compared to best pruning baselines - with LLM-Judge score of $8.87$ vs $7.32$ on physical reasoning on a scale of $0-10$. It further sustains high reasoning consistency (LLM-J $>$ $7.0$) up to $50\%$ pruning, a $20-25$ percentage-point improvement over existing methods. Notably, existing pruning methods suffer from abrupt performance collapse beyond $25-30\%$ pruning, whereas our method maintains high logical validity under LLM-as-judge evaluation and stable semantic similarity even up to $50\%$ pruning levels, while exhibiting comparably lower degradation in perplexity.
\end{abstract}

\section{Introduction}
\label{sec:intro}

In recent years, we have witnessed massive advances in multimodal reasoning through efficient vision and language modeling with systems such as, Qwen2.5-VL \citep{qwen2.5vl}, Gemma-3 \citep{gemma3}, and LLaVA-OneVision \citep{llava-onevision}; which integrate heavily pretrained vision encoders with powerful (large) language decoders. Augmented by techniques like Chain-of-Thought prompting \citep{wei2022chain, kojima2022zeroshotcot}, these models now exhibit strong performance across demanding applications; including medical visual question answering, autonomous driving, and scientific diagram reasoning, where structured, step-by-step inference is essential.
A critical enabler of this performance is chain-of-thought (CoT) prompting \citep{wei2022chain, kojima2022zeroshotcot}, which elicits explicit intermediate reasoning steps before a final answer. 
Following the success in text-based reasoning \citet{wei2022chain}, various works adapted CoT to the multimodal context using text-only CoT \cite{zhang2023multimodal,yang2023mm} and CoT with visual context (bounding boxes \cite{shao2024visual} or visual tokens \cite{chen2025mint}).  However, since large(r) models benefit from CoT more \citep{wei2022chain}, this comes with a substantial computational burden, posing significant challenges for real-world deployment.
Although model pruning offers a promising route toward efficiency, existing methods largely overlook the inherently structured, multi-step nature of chain-of-thought reasoning in multimodal contexts, leading to brittle performance under compression.

Therefore, our central objective in this work is to preserve multimodal CoT reasoning abilities of VLMs under extreme compression.  We specifically adopt the text-only variant as \emph{multimodal chain-of-thought (CoT) reasoning}, representing the process of  answering questions about an image through a sequence of intermediate \emph{text-based} reasoning steps grounded in both visual and textual inputs. To investigate \textit{reasoning} skills, we particularly focus on visual reasoning (or question-answering) benchmarks which require such multi-step reasoning to solve.  For example, questions such as \emph{``If the glass near the table edge is bumped, will water spill onto the laptop below?''} require identifying spatial relationships, applying physical reasoning, and chaining these observations into a causal conclusion.

This problem is challenging and under-explored for two key reasons. \emph{First}, current pruning methods do not explicitly cater to preserving \textit{reasoning coherence} in COT. Existing methods such as Wanda \citep{wanda}, SparseGPT \citep{frantar2023sparsegpt}, and LLM-Pruner \citep{llmpruner} estimate importance via aggregate loss, treating all tokens equally. In long CoT responses, only a small subset of tokens correspond to reasoning transitions, and their contribution is diluted under such aggregation, leading to a signal-to-noise imbalance. \emph{Second}, VLMs introduce additional complexity beyond LLMs due to cross-modal entanglement, architectural heterogeneity, and large disparities in unit granularity across components. As a result, pruning decisions must account not only for reasoning structure but also for preserving vision-language alignment. Motivated by these challenges, we investigate three questions: \textbf{(RQ1)} Can structural pruning preserve the logical coherence of multi-step CoT reasoning in VLMs? \textbf{(RQ2)} Why do existing pruning methods fail to preserve reasoning when applied to VLMs? \textbf{(RQ3)} How should pruning be redesigned to preserve reasoning trajectories under strict parameter budgets?

\paragraph{} Overall, we make the following contributions:
\begin{enumerate}[nosep, leftmargin=*]

    \item \textbf{Failure of traditional pruning methods in VLMs.}
    We show that structured pruning methods (e.g., FLAP, LLM-Pruner, Attribution Pruning~\citep{an2024flap,llmpruner,yang2023attribution}) fail to preserve chain-of-thought (CoT) reasoning in VLMs despite sometimes maintaining acceptable perplexity. These methods treat all CoT tokens uniformly and ignore cross-modal interactions, causing reasoning chains to collapse under moderate compression. 
    
    \item \textbf{A reasoning-aware, training-free pruning method for VLMs.} We introduce \method{}--a \underline{mu}ltimodal \underline{C}oT \underline{r}easoning \underline{a}ware (training-free) \underline{p}runing method, that is targeted to preserve chain-of-thought reasoning quality in VLMs under high structural compression. \method{} integrates reasoning-aware attribution to capture transition-critical (\textit{anchor}) tokens in CoT, cross-modal dependency modeling to retain vision--language alignment, and layer-wise sensitivity to guide structured pruning under a global parameter budget.

    \item \textbf{Extensive benchmarking and analysis for the preservation of reasoning quality.} At different levels of pruning, we benchmark the performance of our method against four pruning baselines on four VLMs across three datasets. We estimate \emph{logical validity} of chain-of-thought (CoT) using an LLM-as-a-judge metric \citep{llmasjudge, chen2024humansllmsjudgestudy}, and report other final answer-level metrics.
    Furthermore, our extensive analysis involving MLP zero-out, \textit{anchor} token ablation, and distributional differences between next token prediction abilities before and after pruning pinpoints how each component of our method supports CoT quality preservation during high pruning levels. 
\end{enumerate}

\section{Related Work}
\label{sec:related}

\textbf{Structured vs.\ Unstructured pruning.}\quad
Neural network pruning research has a long history, spanning early approaches such as optimal brain damage \citep{lecun1989optimal}, optimal brain surgeon \citep{hassibi1993optimal} to modern large-scale methods. Among these, unstructured methods operate by zeroing individual weights. SparseGPT \citep{frantar2023sparsegpt} solves layer-wise sparse regression to achieve $60\%$ sparsity on $175 B$ models with negligible accuracy loss; Wanda \citep{wanda} uses the product of weight and activation magnitude for efficient one-shot pruning. However, this may result in irregular sparsity patterns, which require specialized hardware for inference speedup. In contrast, structured methods remove entire units (neurons, heads, layers), producing dense sub-networks with direct hardware acceleration. LLM-Pruner \citep{llmpruner} uses Taylor-based dependency graphs; SliceGPT \citep{ashkboos2024slicegpt} projects weight matrices to lower dimensions; and FLAP \citep{an2024flap} measures activation fluctuation.

\textbf{Layer-wise vs.\ Global pruning.}\quad
Layer-wise methods, including SparseGPT and Wanda, determine per-layer masks by minimizing local reconstruction error independently. This is computationally convenient, but locally optimal decisions need not compose globally. Global methods \citep{molchanov2017pruning, molchanov2019importance} rank all units across the network by a single importance score. ECoFLaP \citep{ecoflap} bridges the two via zeroth-order global importance for per-layer sparsity ratios. OWL \citep{owl} uses outlier distributions for non-uniform sparsity. We adopt global pruning because VLM structural unit costs span orders of magnitude (${\sim}5.2$M per GQA group vs.\ ${\sim}21$K per MLP neuron), and a global knapsack formulation \citep{martello1990knapsack} naturally handles this heterogeneity.

\textbf{VLM Compression.}\quad
Previous work on VLM pruning is somewhat limited and focuses mainly on token-level pruning of visual inputs (e.g., ATP-LLaVA \citep{atp-llava}, IVTP \citep{ivtp}) or intrinsic dimension-based methods \citep{intrinsicdimension}. These methods operate at the input or activation level without structurally pruning parameters. Among parameter-level approaches, ECoFLaP \citep{ecoflap} handles BLIP-style architectures for captioning and simple VQA. TAMP \citep{tamp} introduces token-adaptive layer-wise sparsity.  Complementary to compression research, recent work \citep{dutta2025nlki} has also improved VLM reasoning through external knowledge integration, demonstrating that lightweight models can benefit substantially from structured knowledge. To the best of our knowledge, no prior work studies structured pruning of VLMs with the explicit objective of preserving chain-of-thought reasoning.

\textbf{Chain-of-Thought (CoT) reasoning.}\quad
The simple technique of CoT prompting \citep{wei2022chain,kojima2022zeroshotcot} has shown drastic improvements in the reasoning ability of LLMs. Following the success, researchers adopted CoT in multimodal setting \citep{zhang2023multimodal_cot}, where the CoT grounds rationales in visual evidence \citep{lu2022scienceqa}. To improve latency and token efficiency, researchers have worked on reducing redundancy \cite{lyu-etal-2023-faithful}, 
preference-based CoT optimization \citep{zhang2024cpo}, and attention-head specialization \citep{voita2019analyzing, michel2019heads}. Knowledge distillation approaches for compressing reasoning capabilities \citep{hinton2015distilling, hsieh2023distilling} are complementary to our pruning framework and can serve as a post-pruning recovery stage. Prior work \citep{aditya2018explicitreasoningendtoendneural, aditya2019spatial} has also explored explicit and interpretable reasoning mechanisms for vision-language systems, including symbolic reasoning layers for VQA and spatial knowledge distillation frameworks that transfer structured reasoning capabilities to neural models. However, no prior work explicitly studies how pruning affects the structural integrity of generated reasoning chains. 

\section{Problem Setup: Structured Pruning for Multimodal COT Reasoning}
\label{sec:background}

We consider the CoT reasoning task in Vision-Language Model (VLM), where a VLM is explicitly prompted to answer a question about an image in a step-by-step fashion. Specifically, given an image and a question as input, we use a CoT prompt \cite{wei2022chain} to instruct the VLM ($M = (\mathcal{E}_{v}, \mathcal{D}_\ell)$) to generate a chain of intermediate reasoning steps before producing a final answer. Intermediate steps, as well as the final answer are sequentially sampled from the model conditioned on previous steps and the input. 

\textbf{Structured Pruning of VLMs.}\quad For pruning, we assume the VLM $M = (\mathcal{E}_{v}, \mathcal{D}_\ell)$ consists of a vision encoder $\mathcal{E}_{v}$ with $l_v$ transformer blocks and a language decoder $\mathcal{D}_\ell$ with $l_\ell$ transformer layers. Each module contains \emph{structural units} ($\mathcal{U}_\ell$): individual neurons for SwiGLU MLPs (a gated MLP variant used in modern transformers) \citep{shazeer2020gluvariantsimprovetransformer}, attention heads for vision blocks, and GQA groups \citep{ainslie2023gqa} for language attention (for more details, refer Appendix). 

We further assume that the importance of the task-based individual network parameter is calculated using the standard First-order Taylor expansion \citep{molchanov2017pruning, molchanov2019importance}. Here, for a structural unit $u$ with weight vector $\mathbf{w}_u$, importance of the unit is approximated as the change in loss upon removal of the unit, given by:
$\Delta \mathcal{L} \approx \mathbf{w}_u^\top \nabla_{\mathbf{w}_u} \mathcal{L}.$ 
We take absolute values to ensure non-negative importance, and aggregate over a calibration dataset $\mathcal{D}$ to yield a global importance score. This forms the backbone of structured pruning methods for traditional neural networks and LLMs \citep{llmpruner}, as well as our approach.

\section{Our Method: \method{}}

\label{sec:method}
Given a VLM ${M}$ and a target pruning ratio $S$, our goal is to identify and remove structural units, that contribute least to chain-of-thought reasoning quality, while respecting a global parameter budget. Our algorithm operationalizes two hypotheses: 1) we \textit{posit} that there are a few important transition regions (marked by \textit{anchor} tokens) in CoT which are crucial for governing reasoning coherence, and 2) a VLM-aware pruning method must take the activation distribution difference between vision and language components. 

\begin{figure}[t]
\centering
\includegraphics[width=\linewidth]{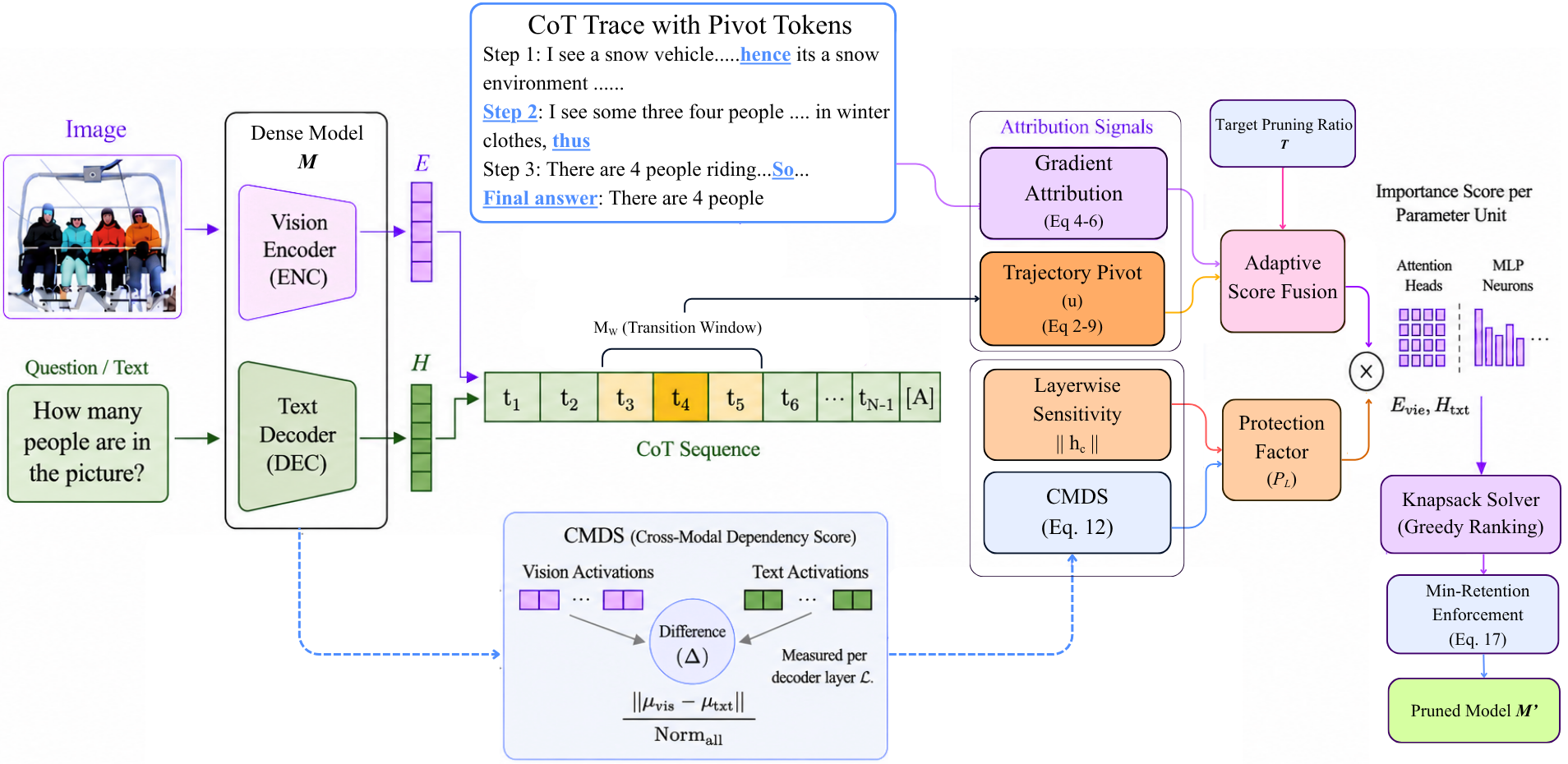}
\caption{\textbf{Overview of \textsc{\method{}}.} Gradient attribution and trajectory-pivot attribution provide complementary pruning signals, which are fused using a pruning-ratio-dependent coefficient $\gamma_{\text{dyn}}$. Per-layer cross-modal dependency (CMDS) and output sensitivity define a protection factor $p_f$ that preserves vision-language integration layers under aggressive pruning. The resulting importance scores are used by a greedy knapsack solver to allocate the global pruning budget across attention heads and MLP neurons while enforcing minimum retention constraints.}
\label{fig:method_diagram}
\end{figure}

\textbf{Algorithm.}~We define our method $\mu$CRASP as a sequence of four principal steps (Fig~\ref{fig:method_diagram}). 

\noindent\underline{First}, for each structural unit, we compute a standard global attribution score $I_{\text{global}}$ via first-order Taylor expansion over all token positions (Appendix\ref{sec:global_attr}). 

\underline{Second}, we compute trajectory pivot attribution score $A_{\text{pivot}}$ by restricting attribution to local windows around detected transition regions (Appendix\ref{sec:tpa}). These transition regions are, in turn detected by identifying a few \textit{pivot} (or reasoning transition) \textit{tokens}, which occur near boundaries between successive reasoning steps. These tokens correspond to points where the model consolidates the conclusion of one intermediate step and initiates the next. For example, in a multi-step visual reasoning task, a transition may occur when the model shifts from describing observed entities (e.g., identifying objects or spatial relations) to making an inference or causal conclusion. Empirically, such transitions are sparse ($\approx5$\% of tokens) yet disproportionately important, as perturbing them disrupts logical continuity. To approximate these transition points, we apply a lightweight detection strategy that identifies two classes of surface markers: 
\emph{structural delimiters}, such as numbered step prefixes (``Step 1:'', ``2.'') and terminal markers (``Final Answer:'', ``Conclusion:'', ); and \emph{logical connectives}, such as causal or conclusory conjunctions (``Therefore'', ``Thus'', ``So'', ``Hence'', ``This means'', ``and'' etc). When neither classes of markers is present, we resort to considering tokens at sentential boundaries. Despite the heuristic, our overall method is resistant to variations in choice of structural patterns, as i) we depend on conventional linguistic and structural cues, ii) we define a window of $W$ tokens around the markers, and iii)  the rest of our method ensures a dependency on the transition \textit{region(s)} than specific tokens (later validated by experiments in Appendix~\ref{app:random_pivots}).

\underline{Third}, we quantify cross-modal affinity of each decoder layer using a \emph{Cross-Modal Dependency Score} (CMDS) --a measure inspired by Maximum Mean Discrepancy (MMD)~\citep{gretton2012kernel}. MMD is a kernel-based statistical distance between probability distributions; motivated by this formulation, we formulate it to quantify the discrepancy between activation distributions induced by visual and textual tokens within a layer. This allows CMDS to capture the extent to which a layer contributes to cross-modal interaction, rather than merely measuring distributional differences in a generic sense (Appendix\ref{sec:cmds}). We additionally measure structural load via output activation magnitude (Appendix\ref{sec:sensitivity}). 

\underline{Fourth}, we fuse $I_{\text{global}}$ and $A_{\text{pivot}}$ with a compression-conditioned mixing coefficient $\gamma_{\text{dyn}}$ that progressively decreases the reasoning-specific signal as the parameter budget tightens. Here, we additionally factor in individual layer-wise activation distribution (aka \textit{sensitivity}) as well as CMDS scores. Therefore, we scale the resulting importance by layer-level protection factors derived from CMDS and \textit{sensitivity} (Appendix\ref{sec:knapsack}). 
 
\noindent\textbf{Objective.} Lastly, we solve a global knapsack problem over all structural units, ranking them by value-to-cost efficiency and selecting units greedily until the parameter budget is met, with minimum-retention constraints to prevent layer collapse. This formulation is necessary due to the extreme heterogeneity of structural units in VLMs: for example, a single GQA group in Qwen2.5-VL-7B contains $\sim$5.2M parameters, whereas an MLP neuron contains only $\sim$21K. Uniform layer-wise pruning leads to disproportionate removal of low-cost units while retaining expensive ones. By contrast, global allocation optimizes the trade-off between importance and parameter cost across all units. 
Given the target pruning ratio $S \in (0,1)$, we retain a subset of structural units (a keep-set $\mathcal{K} \subseteq \bigcup_\ell \mathcal{U}_\ell$), which maximizes reasoning-preserving importance by satisfying:
\begin{equation}
\label{eq:main_objective}
\max_{\mathcal{K} \subseteq \bigcup_\ell \mathcal{U}_\ell} \;\sum_{u \in \mathcal{K}} I(u)
\quad \text{subject to} \quad 
\sum_{u \in \mathcal{K}} c(u) \leq (1 - S)\sum_{u \in \bigcup_\ell \mathcal{U}_\ell} c(u),
\end{equation}
where $I(u)$ denotes the (reasoning-aware) importance of unit $u$, $c(u)$ is the parameter count of unit $u$. We provide more details of the four components: global attribution  in Appendix\ref{sec:global_attr}, trajectory pivot attribution in Appendix\ref{sec:tpa}), layer-level profiling via CMDS and sensitivity in Appendix\ref{sec:cmds} \& \ref{sec:sensitivity}), and dynamic fusion with global knapsack allocation  in Appendix\ref{sec:knapsack}). Fig~\ref{fig:method_diagram} provides an overview and Algorithm~\ref{alg:vcot_prune} gives the complete procedure.


\section{Experimental Setup}
\label{sec:experiments}

We design experiments to address three questions: \emph{Can} structural pruning preserve CoT reasoning in VLMs (RQ1)? \emph{Why} do existing pruning methods fail at this task (RQ2)? And \emph{how} does reasoning-aware pruning overcome these failures (RQ3)? Each design decision-choice of metrics, benchmarks, architectures, and baselines-targets one or more of these questions.

\textbf{Architectures: spanning model families and scales.}\quad
We evaluate four VLMs that span three architectural families and a large parameter range:  \textit{LLama 3.2-$11B$}~\citep{llama3herdmodels}  ($10.1B$, $40$ layers), \textit{Qwen2.5-VL-$7B$}~\citep{qwen2.5vl} ($7.9B$, $28$ layers), \textit{Gemma-3-$4B$}~\citep{gemma3} (4.3B, 34 layers), \textit{Qwen2.5-VL-$3B$} (3.8B, $36$ layers), and \textit{Qwen2-VL-2B}~\citep{qwen2vl} ($2.2B$, $28$ layers). The Qwen family uses a ViT-based encoder with patch merging, while Gemma uses SigLIP with pooling-based projection, spanning various VL fusion architectures. All models employ grouped-query attention (GQA) (Appendix Table ~\ref{tab:architecture}).

\textbf{Datasets: testing visual grounding and reasoning across domains.}\quad
We select three datasets, each concentrating on a different facet of visually grounded multi-step reasoning and ensuring that our evaluation is not dominated by a single reasoning type:
(1)~\textit{TDIUC-Physical}~\citep{kafle2017tdiuc} requires reasoning about spatial layout, colour, shape, and material properties - tasks where the model must ground each reasoning step in specific visual features.
(2)~\textit{TDIUC-Quantitative}~\citep{kafle2017tdiuc} requires counting, numerical comparison, and arithmetic inference, testing whether pruning disrupts the precise token-level operations needed for numerical answers.
(3)~\textit{A-OKVQA}~\citep{schwenk2022aokvqa} requires commonsense reasoning grounded in visual scenes (25K questions over COCO), testing whether pruning destroys the broadly distributed world knowledge that supports commonsense inference.
For each dataset, we generate CoT  traces using GPT-4o, validate using manual evaluation and consider it as ground-truth trace. We use the held-out test sets for evaluation. For further details, see Appendix~\ref{app:datasets} \& \ref{app:cot_generation}.


\textbf{Metrics: evaluating reasoning, not just token accuracy.}\quad
Prior pruning work evaluates compressed models almost exclusively via perplexity, BLEU, F1, or direct-answer accuracy~\citep{llmpruner,an2024flap,frantar2023sparsegpt}. These metrics capture surface-level fluency and lexical overlap, but do not quantify logical validity of reasoning steps after pruning. 
To address this gap, we introduce \textbf{LLM-as-Judge (LLM-J)} as one of our key metric. Here, we use \textsc{GPT-3.5-Turbo} to evaluate the logical validity of each generated CoT trace against the ground-truth chain and assigns a score on a 0-10 scale\footnote{In Appendix~\ref{app:llm_judge}, we justify the choice of the scale with a comparative analysis.}(Appendix\ref{app:llm_judge}). 

We complement LLM-J with six additional metrics that capture distinct failure modes.
\textbf{Perplexity (PPL)} measures token-level prediction confidence. 
\textbf{Exact Match ($EM_{a}$)} and \textbf{Semantic similarity ($SEM_{c+a}$)} test of the CoT + final answer precision respectively. \textbf{Formatting Score} verifies structural integrity of the CoT trace; and \textbf{BLEU} and \textbf{F1} provide standard surface-level baselines. These metrics help us disentangle reasoning collapse from answer-extraction failure. 

\textbf{Pruning Baselines \& Calibration Protocol.}\quad
We construct a calibration set by randomly sampling $128$ image, question pairs from the training split and pairing them with synthetic CoT traces generated by GPT-4o (Appendix~\ref{app:cot_generation}). We use the same calibration set for all the baselines and our method.

We compare against five structured pruning methods, all using our calibration set: (1)~\textbf{Magnitude}: removes units with smallest aggregate $\ell_1$ weight norm~\citep{zhu2017prune}. (2)~\textbf{Attribution  Pruning}: task-specific gradient-based saliency~\citep{yang2023attribution}, providing a stronger task-aware baseline. (3)~\textbf{LLM-Pruner}: dependency-graph-based group pruning with first and second-order Taylor importance~\citep{llmpruner}. (4)~\textbf{Wanda-SP}: the structured variant of Wanda~\citep{wanda} as implemented by~\citep{an2024flap}, using the product of weight magnitude and input activation norm. (5)~\textbf{FLAP}: fluctuation-based adaptive structured pruning~\citep{an2024flap}. We provide further details of the baselines in Appendix\ref{app:baselines}.

\begin{figure}[!htb]
\centering
\includegraphics[width=\linewidth]{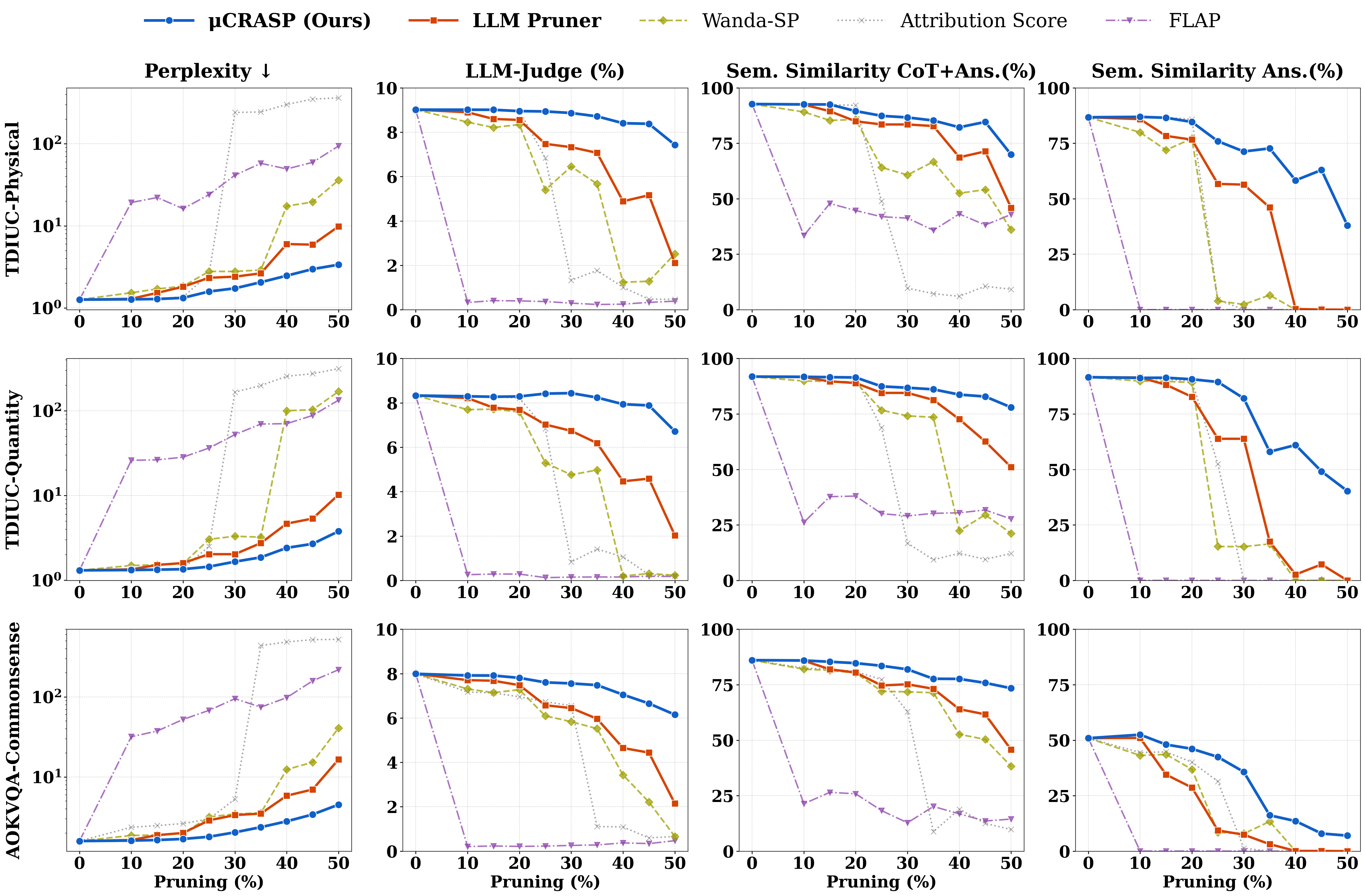}
\label{fig:main_experiment}
\caption{Performance across pruning methods under increasing pruning level on Qwen2.5-VL-7B across three reasoning domains. Each row corresponds to a domain; columns report Perplexity, LLM-Judge, Semantic Similarity (cosine) of CoT + final answer, and Semantic similarity (semantic) of final answer. \emph{We report Magnitude Pruning \citep{layeradaptivesparsitymagnitudebasedpruning} results in Appendix~\ref{sec:appendix_magnitude}, due its low performance.}}
\end{figure}

\section{Experimental Results: Compression vs. Performance}
\label{sec:results}
 We report the compression vs./ performance curves for Qwen2.5-VL-7B in Fig.2 and cross-model results at 30\% pruning across all five VLMs in Table~\ref{tab:main_all_models}. Full compression--performance curves for all the VLMs with additional token level metrics are provided in Appendix~\ref{sec:appendix_full_curves}.


\begin{table}[t]
\centering 
\caption{\method{} versus baselines at \textbf{$30\%$ pruning ratio} across five VLMs and three reasoning domains. We report LLM-Judge (\textit{J}), String match of final Answer (\textit{$EM_{a}$}), Semantic Similarity (cosine) of CoT + final Answer (\textit{$SEM_{c+a}$}), Perplexity (\textit{PPL}) and the unpruned model performance \textit{Dense M.} We have put \textit{INF} where the PPL value is more than $1K$.} 
\label{tab:main_all_models}
\small
\setlength{\tabcolsep}{3pt}
\resizebox{\columnwidth}{!}{%
\begin{tabular}{ll rrrr rrrr rrrr}
\toprule
& & \multicolumn{4}{c}{\textbf{Physical}} & \multicolumn{4}{c}{\textbf{Quantitative}} & \multicolumn{4}{c}{\textbf{Commonsense}} \\
\cmidrule(lr){3-6} \cmidrule(lr){7-10} \cmidrule(lr){11-14}
\textbf{\#} & \textbf{Method} 
& J & $EM_{a}$ & $SEM_{c+a}$ & PPL
& J & $EM_{a}$ & $SEM_{c+a}$ & PPL
& J & $EM_{a}$ & $SEM_{c+a}$ & PPL \\
\midrule

\multirow{4}{*}{\rotatebox{90}{\footnotesize Llama 11B}}

& \textbf{Dense M.} & \textbf{8.85} & \textbf{85.8} & \textbf{91.82} & \textbf{1.44} & \textbf{7.91} & \textbf{89.3} & \textbf{90.88} & \textbf{1.44} & \textbf{7.8} & \textbf{51.7} & \textbf{85.6} & \textbf{1.72} \\
\cmidrule{2-14}

& \textbf{\method{}} & \textbf{8.55} & \textbf{80.7} & \textbf{89.72} & \textbf{1.80} & \textbf{7.29} & \textbf{79.8} & \textbf{87.01} & \textbf{1.88} & \textbf{6.66} & \textbf{34.9} & \textbf{80.41} & \textbf{2.39} \\

& LLM-Pruner & 4.69 & 0.0 & 65.62 & 3.57 & 2.44 & 0.0 & 54.46 & 3.89 & 2.77 & 0.0 & 55.41 & 4.96 \\
& Wanda-SP & 5.0 & 0.0 & 60.45 & 3.21 & 3.02 & 0.0 & 57.46 & 3.60 & 4.43 & 0.0 & 57.95 & 4.63 \\
& Attribution P. & 8.07 & 0.0 & 85.61 & 2.41 & 5.78 & 9.46 & 76.77 & 2.49 & 6.95 & 23.93 & 79.99 & 2.81 \\
& FLAP & 2.79 & 0.0 & 56.33 & 5.98 & 3.41 & 0.0 & 47.66 & 7.18 & 2.02 & 0.0 & 44.90 & 12.2 \\
\midrule
\multirow{4}{*}{\rotatebox{90}{\footnotesize Qwen 7B}}

& \textbf{Dense M.} & \textbf{9.01} & \textbf{86.73} & \textbf{92.69} & \textbf{1.26} & \textbf{8.33} & \textbf{91.6} & \textbf{91.85} & \textbf{1.31} & \textbf{8.0} & \textbf{50.9} & \textbf{86.07} & \textbf{1.58} \\
\cmidrule{2-14}

& \textbf{\method{}} & \textbf{8.87} & \textbf{71.3} & \textbf{86.64} & \textbf{1.74} & \textbf{8.44} & \textbf{82.1} & \textbf{86.84} & \textbf{1.66} & \textbf{7.56} & \textbf{35.72} & \textbf{81.93} & \textbf{2.0} \\

& LLM-Pruner & 7.32 & 56.4 & 83.48 & 2.41 & 6.74 & 63.8 & 84.51 & 2.03 & 6.45 & 7.4 & 75.19 & 3.4 \\
& Wanda-SP & 6.45 & 2.4 & 60.74 & 2.79 & 4.76 & 15.2 & 74.15 & 3.31 & 5.83 & 7.9 & 71.78 & 3.5 \\
& Attribution P. & 1.33 & 0.0 & 9.72 & 241 & 0.08 & 0.0 & 16.78 & 166.6 & 6.56 & 1.31 & 62.91 & 5.3 \\
& FLAP & 0.3 & 0.0 & 41.28 & 41.47 & 0.02 & 0.0 & 29.07 & 52.49 & 0.2 & 0.0 & 12.78 & 95.1 \\
\midrule

\multirow{4}{*}{\rotatebox{90}{\footnotesize Gemma 4B}}
& \textbf{Dense M.} & \textbf{8.65} & \textbf{83.53} & \textbf{92.06} & \textbf{1.28} & \textbf{7.85} & \textbf{90.23} & \textbf{91.08} & \textbf{1.38} & \textbf{7.02} & \textbf{46.55} & \textbf{84.31} & \textbf{1.71} \\
\cmidrule{2-14}

& \textbf{\method{}} & \textbf{8.02} & \textbf{77.3} & \textbf{88.71} & \textbf{1.9} & \textbf{6.95} & \textbf{87.4} & \textbf{87.89} & \textbf{2.1} & \textbf{6.26} & \textbf{31.7} & \textbf{80.03} & \textbf{2.4} \\
& LLM-Pruner & 0.78 & 0.1 & 40.37 & 11.9 & 0.45 & 0.0 & 42.74 & 10.5 & 0.67 & 0.0 & 38.67 & 10.4 \\
& Wanda-SP & 1.92 & 0.1 & 52.85 & 3.5 & 1.61 & 1.3 & 62.56 & 3.51 & 1.29 & 0.7 & 54.83 & 4.9 \\
& Attribution P. & 7.72 & 73.4 & 86.18 & 2.12 & 6.94 & 83.8 & 85.55 & 2.3 & 6.19 & 28.4 & 78.57 & 2.4 \\
& FLAP & 0.3 & 0.0 & 35.06 & 17.7 & 0.04 & 0.0 & 25.76 & 49.87 & 0.09 & 0.0 & 8.89 & 44.2 \\
\midrule 

& \textbf{Dense M.} & \textbf{8.86} & \textbf{85.97} & \textbf{92.08} & \textbf{1.36} & \textbf{8.33} & \textbf{91.6} & \textbf{91.85} & \textbf{1.31} & \textbf{7.50} & \textbf{49.08} & \textbf{84.95} & \textbf{1.74} \\
\cmidrule{2-14}

\multirow{4}{*}{\rotatebox{90}{\footnotesize Qwen 3B}}
& \textbf{\method{}} & \textbf{8.07} & \textbf{24.8} & \textbf{80.64} & \textbf{2.57} & \textbf{7.51} & \textbf{41.0} & \textbf{83.79} & \textbf{2.35} & \textbf{6.70} & \textbf{23.1} & \textbf{77.54} & \textbf{3.4} \\
& LLM-Pruner & 6.73 & 0.1 & 64.61 & 3.49 & 5.64 & 0.0 & 60.35 & 3.12 & 5.11 & 0.0 & 62.78 & 3.5 \\
& Wanda-SP & 5.34 & 0.0 & 59.96 & 7.47 & 0.5 & 0.0 & 53.77 & 21.27 & 3.63 & 0.2 & 50.36 & 18.9 \\
& Attribution P. & 0.01 & 0.0 & 3.77 & \textit{INF} & 0.01 & 0.0 & 5.05 & \textit{INF} & 0.02 & 0.0 & 8.80 & \textit{INF} \\
& FLAP & 0.03 & 0.0 & 32.49 & 66.9 & 0.06 & 0.0 & 38.09 & 40.43 & 0.05 & 0.0 & 27.76 & 99.2\\
\midrule 

& \textbf{Dense M.} & \textbf{8.39} & \textbf{78.37} & \textbf{90.04} & \textbf{1.76} & \textbf{7.55} & \textbf{87.45} & \textbf{89.35} & \textbf{1.8} & \textbf{6.97} & \textbf{39.65} & \textbf{81.98} & \textbf{2.21} \\
\cmidrule{2-14}

\multirow{4}{*}{\rotatebox{90}{\footnotesize Qwen 2B}}
& \textbf{\method{}} & \textbf{7.65} & \textbf{63.4} & \textbf{86.98} & \textbf{2.4} & \textbf{6.65} & \textbf{52.0} & \textbf{83.55} & \textbf{2.37} & \textbf{5.3} & \textbf{13.9} & \textbf{76.02} & \textbf{3.8} \\
& LLM-Pruner & 5.35 & 0.0 & 72.57 & 4.6 & 4.39 & 0.6 & 68.76 & 3.84 & 4.91 & 0.4 & 69.10 & 4.2 \\
& Wanda-SP & 0.29 & 0.0 & 25.60 & 15.6 & 0.9 & 0.0 & 18.90 & 12.92 & 0.6 & 0.0 & 10.36 & 17.4 \\
& Attribution P. & 7.17 & 17.2 & 81.17 & 2.7 & 6.26 & 13 & 74.84 & 2.71 & 5.21 & 5.9 & 72.57 & 3.5 \\
& FLAP & 0.9 & 0.0 & 39.76 & 15.5 & 0.7 & 0.0 & 35.08 & 17.7 & 1.13 & 0.0 & 34.80 & 20.7 \\
\bottomrule
\end{tabular}%
}
\end{table}

\footnotetext{In Table~\ref{tab:main_all_models} $EM_a$ is computed using strict token-level string matching between the predicted and ground-truth final answer. As a result, semantically correct but syntactically verbose responses receive zero score despite conveying the correct answer. we therefore additionally report Semantic Similarity ($SEM_{c+a}$) in table and other token level metrics in the model compression-performance curves in Appendix~\ref{sec:appendix_full_curves}.}

We organize the findings around three themes that hold \emph{in all models}.

\textbf{\method{} preserves reasoning coherence while exact-match accuracy degrades first.}\quad 

We observe that across all four VLMs, LLM-J and $EM_{a}$ follow \textit{visibly} different degradation trajectories under \method{}. LLM-J degrades minimally even at high pruning ratio, while $EM_{a}$ collapses much earlier and more steeply. In Qwen2.5-VL-7B (Physical), the unpruned model receives an LLM-J score of $9.02$ and $EM_{a}$ of $86.7$. At $30\%$ pruning, LLM-J drops by only $0.15$ points to $8.87$, while $EM_{a}$ drops by $15.4$ points to $71.3$. Similar patterns can also be observed for Llama, where LLM-J drop from $8.85$ to $8.55$ (only by $0.3\%$), and $EM_{a}$ from $85.83$ to $80.77$ (dropped by $5.06\%$). The asymmetry is more stark on \underline{smaller} models: Qwen2.5-VL-3B at $30\%$ retains LLM-J $8.07$ ($-9\%$ from dense $8.86$), but $EM_{a}$ collapses from $86.0$ to $24.8$ ($-71\%$); similarly, Qwen2-VL-2B at $40\%$ pruning holds LLM-J at $6.93$ while $EM_{a}$ falls to $14.0$. 
Interestingly, we observe this behavior distinctively for \method{}, and not other pruning methods. For example, when Gemma-3-4B's reasoning collapses under LLM-Pruner beyond $20\%$ pruning ratio (LLM-J: $6.77 \to 0.54$ at $25\%$), $EM_{a}$ drops simultaneously from $50.8$ to $0.03$. Likewise, Qwen2.5-VL-3B under FLAP collapses at just $10\%$ pruning ratio (LLM-J $0.37$, $EM_{a}$ $0.03$), with no separation between the two metrics. This indicates that, \method{} treats reasoning coherence as a \emph{separable} capability, preserving logical chain structure even as the verbatim precision of the final answer degrades.

\textbf{Scaling behaviour across model sizes.}\quad
We compare \method{} against the best available baseline on the largest and smallest models at $30\%$ pruning in Table~\ref{tab:main_all_models}. On Qwen2.5-VL-7B, \method{} improves LLM-J by $+1.54$, $+1.70$, and $+1.0$ points on Physical, Quantitative, and Commonsense respectively, while simultaneously recovering $+14.9$, $+18.2$, and $+34.4$ points of exact-match accuracy ($EM_{a}$). On the $2$B model, LLM-J gains are more modest ($+0.48$, $+0.39$ points on Physical and Quantitative), but the $EM_{a}$ improvements are $+46.2$ and $+39.0$ points, indicating that \method{} preserves answer-extraction capability even when reasoning coherence is partially degraded by compression. These results confirm that \method{}'s gains are robust across model scales. For smaller models, \method{} achieves consistently large $EM_{a}$ gains with moderate LLM-J improvements. Qualitative example of generated reasoning traces for our method and baselines at $50\%$ pruning are shown in Appendix~\ref{app:qualitative}.
 
\textbf{Domain-specific degradation patterns are consistent across model scales.}\quad
We observe that Physical reasoning degrades the slowest. Here, all five models retain LLM-J within $10\%$ of their dense baseline at $30\%$ pruning (Qwen2.5-VL-7B: $8.87$ vs dense $9.02$; Gemma-3-4B: $8.02$ vs./ $8.65$; Qwen2-VL-2B: $7.65$ vs./ $8.39$).  

For quantitative reasoning, \method{} preserves reasoning coherence to a large extent while losing numerical precision: LLM-J remains high at $30\%$ (Llama 3.2: $7.29$, Qwen2.5-VL-7B: $8.44$; Gemma-3-4B: $6.95$) while $EM_{a}$ drops sharply beyond this point (Qwen2.5-VL-7B: $82.1 \to 61.0$ at $40\%$; Qwen2-VL-2B: $52.0 \to 2.1$ at $40\%$). Commonsense reasoning declines most uniformly, with LLM-J decreasing at a near-constant rate through $40\%$ pruning on both larger models Qwen2.5-VL-7B: $7.56 \to 7.05$; Gemma-3-4B: $6.26 \to 5.65$ and without the steep declines seen in baselines. Collectively, these experiments show that \method{}'s compression behavior is governed by the structural demands of each reasoning type, not by model or domain-specific factors.

\section{Analysis}
\label{sec:analysis}

\begin{figure}[t]
\centering
\includegraphics[width=\linewidth]{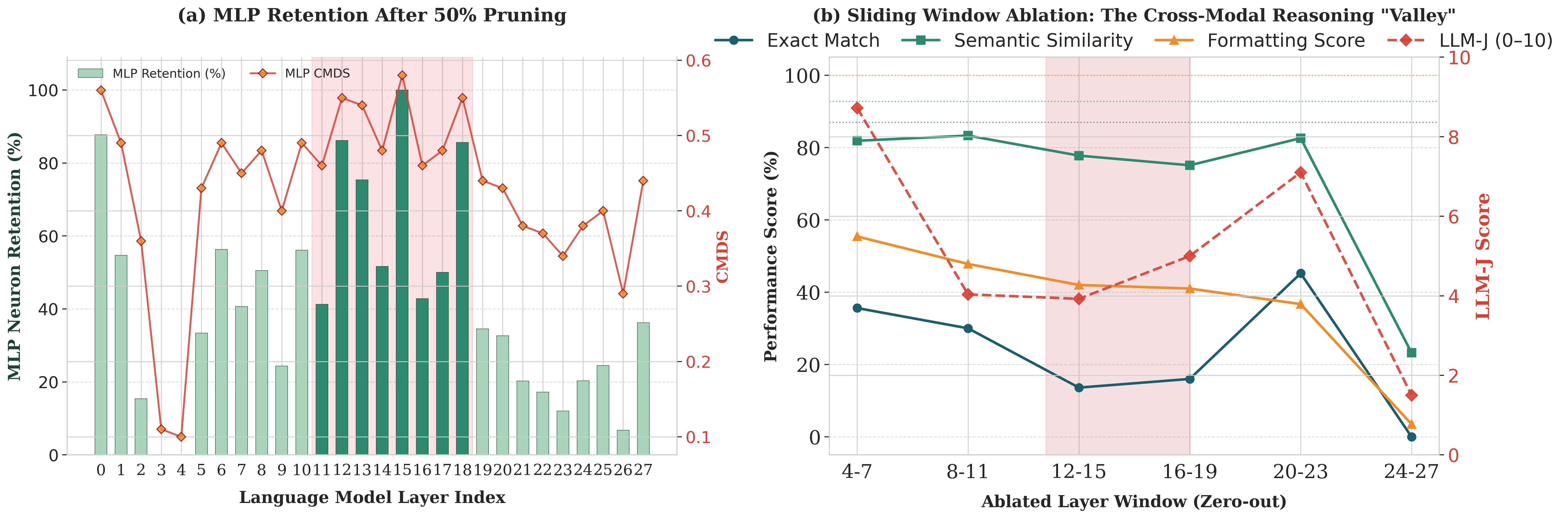}
\caption{(a)~MLP neuron retention after $40\%$ pruning (bars) overlaid with per-layer MLP CMDS (red line). Layers in the peak CMDS region ($11-18$) retain substantially more neurons. (b)~Sliding window ablation on Qwen2.5-VL-7B Physical: zeroing out four contiguous layers at each position. Ablating low-CMDS layers ($4\text{-}7$) causes only moderate degradation, whereas ablating high-CMDS layers ($12\text{-}15$, $16\text{-}19$) triggers catastrophic structural collapse across all metrics, confirming that CMDS correctly identifies the cross-modal bottleneck.}
\label{fig:cmds_plot}
\end{figure}

\textbf{a) CMDS identifies and shields the cross-modal bottleneck.}
In Fig~\ref{fig:cmds_plot}(a), we present the CMDS scores across all $28$ decoder layers of Qwen. For MLPs, CMDS remains low in the early layers ($0.09$--$0.36$ for layers $3$--$7$), rises sharply in the middle layers $11$--$18$ (reaching $0.61$ at layer~$15$ and $0.60$ at layer~$18$), and declines again toward the final layers. This pattern reflects a progression from modality-agnostic token processing in early layers, to vision-language fusion (\textit{grounding}) in the middle, and finally to prediction-focused computation (\textit{reasoning}) in the later stages.
\method{} leverages this signal to allocate its retention budget. Beyond $40\%$ pruning, it preserves $60$--$100\%$ of MLP neurons in the peak-CMDS region, while retaining only $5$--$15\%$ in low-CMDS layers ($1$--$7$) (Fig~\ref{fig:cmds_plot}(a)). In contrast, SOTA pruning baselines ignore this pattern. For instance, Wanda-SP and Attribution Pruning prune the middle layers indiscriminately, disrupting vision-language coupling. On Qwen2.5-VL-7B, Attribution Pruning induces a sharp perplexity increase from $2.57$ to $241$ between $25\%$ and $30\%$ pruning ratio, indicating catastrophic failure due to removing units critical for cross-modal integration. Moreover, our ablations in Appendix~\ref{app:ablations} show that removing CMDS causes sharp performance degradation in our framework.

\textbf{But, does CMDS identify the true bottleneck?}
To verify that CMDS correctly identifies cross-modal bottlenecks, we conduct a layer-wise ablation experiment. Using $500$ random samples from the Physical domain, we zero out the MLP weights of four contiguous layers at five sliding-window positions across the decoder (Figure~\ref{fig:cmds_plot}(b))\footnote{Performing MLP zero out ablation per layer, does not show the degradations properly. Therefore, we group consecutive layers for this experiment.}. Here, initial layers ($4-7$) with low CMDS scores cause graceful degradation: Semantic Similarity ($SEM_{c+a}$) remains high (${\sim}$83\%), Formatting Score stays near $57\%$, and LLM-J holds at ${\sim}$8.8 -- the model still reasons coherently but over corrupted early-stage features. When we zero out layers with high CMDS scores ($12$--$15$ or $16$--$19$), this triggers a qualitatively different failure: the formatting score collapses, semantic similarity crashes, and LLM-J drops sharply. These failure modes confirm that layers $11$--$18$ serve specifically as the structural bottleneck for multimodal reasoning, and pruning them severs the network's ability to integrate visual context into the linguistic reasoning trajectory.
\method{}'s CMDS-based protection dynamically shields precisely this region, which explains why it sustains reasoning quality where baselines that distribute pruning uniformly across layers cannot.

\textbf{b) Pivot attribution protects reasoning-transition tokens.}
At $30\%$ pruning on Qwen2.5-VL-7B Physical, \method{} achieves an LLM-J score of $8.9$ while the strongest remaining baseline, LLM-Pruner, reaches only $7.3$, and Attribution Pruning collapses to $1.3$. The gap between \method{} and LLM-Pruner is not explained by overall token-prediction quality. Both methods operate under the same parameter budget and report similar perplexity what differs is which tokens drive their respective importance scores. Methods that aggregate attribution uniformly across all positions preserve units responsible for fluent next-token prediction while discarding those that maintain logical structure across reasoning steps, because reasoning-transition tokens constitute only ${\sim}5\%$ of the sequence and their signal is diluted under uniform aggregation. Trajectory pivot attribution corrects this by concentrating importance on tokens at step boundaries and reasoning transitions ensuring the keep-set retains units critical for inter-step coherence rather than only for surface-level fluency.

\textbf{c) \method{} preserves the dense model's output distribution.}\quad
We further verify that \method{} preserves the dense model's token-level predictive distribution by measuring per-token KL divergence $D_{\mathrm{KL}}(p_{\text{dense}} \| p_{\text{pruned}})$ over the full vocabulary on $512$ calibration samples at $30\%$ pruning (Fig.~\ref{fig:kl_div}). \method{} achieves the lowest mean KL divergence ($0.92$ nats/token), outperforming the next-best baseline, LLM-Pruner ($1.53$), by $1.7{\times}$ and Attribution Pruning ($10.51$) by $11.4{\times}$. Its KL density remains sharply concentrated near zero across all domains, indicating near-identical logit distributions between dense and pruned models. In contrast, Wanda-SP and FLAP exhibit substantially broader divergence profiles, while Attribution Pruning shows severe bimodal drift, consistent with its catastrophic perplexity at $30\%$ pruning ratio. These trends closely mirror the LLM-J results in Section~\ref{sec:results}, suggesting that \method{} preserves the underlying logit geometry rather than merely surface-level textual similarity. Full distributional analysis with the plot is in Appendix~\ref{app:logit_divergence}.


\textbf{d) Why does \method{} generalize across reasoning domains?}\quad
\method{} delivers consistent LLM-Judge improvements over the strongest baseline (LLM-Pruner) across all three reasoning domains at $30\%$ pruning, with mean gains of $+3.1$ on Physical, $+3.1$ on Quantitative, and $+2.2$ on Commonsense (all on a $0$--$10$ scale), averaged across all four VLMs. The improvement is not concentrated in a single reasoning type or model scale: gains are positive in every one of the twelve model--domain pairs, with the smallest observed delta being $+0.35$ (Qwen2-VL-2B, Commonsense) and the largest $+7.2$ (Gemma-3-4B, Physical, where all baselines collapse). Across all model--domain pairs at $30\%$ pruning, \method{} achieves an overall mean
LLM-J of $7.3$ versus $4.6$ for LLM-Pruner, a $+2.8$ point improvement. The key takeaway is that CMDS-aware pruning and pivot attribution jointly provide a \emph{domain-agnostic} benefit: by protecting the architectural structures responsible for cross-modal integration and reasoning transitions, \method{} preserves CoT quality regardless of which reasoning capability the benchmark exercises.


\section{Conclusion}
\label{sec:conclusion}

We introduced \method{}, a structured pruning framework for VLMs that preserves
chain-of-thought reasoning under high compression.
The core insight is that CoT coherence is governed by sparse pivot tokens at
reasoning-step boundaries whose importance is diluted under the token-uniform
aggregation used by prior methods~\citep{llmpruner,an2024flap,wanda,
frantar2023sparsegpt}. \method{} corrects this by combining pivot-restricted Taylor attribution with a Cross-Modal Dependency Score~\citep{gretton2012kernel} that shields vision--language integration layers from aggressive pruning, two protections absent from prior VLM compression work ~\citep{ecoflap,tamp,intrinsicdimension}. At $50\%$ pruning the gap widens further: \method{} retains a mean LLM-J of $3.90$ while LLM-Pruner falls to $1.56$, demonstrating that \method{} degrades where baselines collapse. Our analysis additionally reveals that reasoning coherence and answer-extraction accuracy degrade along fundamentally different trajectories under \method{}, logical chain structure is preserved long after verbatim answer precision begins to deteriorate; a decoupling that never occurs in any baseline and that motivates richer evaluation protocols for compressed generative models~\citep{lanham2023measuringfaithfulnesschainofthoughtreasoning}. \method{} performs no fine-tuning. And, we leave post-pruning recovery methods such as knowledge distillation~\citep{hinton2015distilling} or parameter-efficient adaptation to future work.

\bibliographystyle{plainnat}
\bibliography{references}

\newpage
\appendix
\renewcommand{\thetable}{\Alph{table}}
\setcounter{table}{0}
\section{Ablation studies}
\subsection{\method{}: Is each component necessary?}
\label{app:ablations}
We systematically ablate each component of \method{} on Qwen2.5-VL-7B at $50\%$ pruning, an aggressive regime that exposes the contribution of each mechanism.
 
\begin{table}[H]
\centering
\caption{Component ablation on Qwen2.5-VL-7B at $50\%$ pruning. Each row removes one component from the full \method{} framework, where \textit{J} stands for LLM-Judge score, \textit{$EM_{a}$} for string exact match of final Answer, \textit{$SEM_{c+a}$} for Semantic  Similarity (cosine) of the complete \textit{CoT + final answer}, \textit{Fmt} for Formatting Score which measures if the model can generate step-by-step formatted reasoning steps}
\label{tab:component_removal_ablation}
\small
\setlength{\tabcolsep}{3pt}
\begin{tabular}{l rrrr rrrr rrrr}
\toprule
& \multicolumn{4}{c}{\textbf{Physical}} & \multicolumn{4}{c}{\textbf{Quantitative}} & \multicolumn{4}{c}{\textbf{Commonsense}} \\
\cmidrule(lr){2-5} \cmidrule(lr){6-9} \cmidrule(lr){10-13}
\textbf{Variant} & J & $EM_{a}$ & $SEM_{c+a}$ & Fmt & J &$EM_{a}$ & $SEM_{c+a}$ & Fmt & J & $EM_{a}$ & $SEM_{c+a}$ & Fmt \\
\midrule
Full \method{} & \textbf{7.43} & \textbf{38.0} & \textbf{69.9} & \textbf{26.5} & \textbf{6.02} & \textbf{23.0} & 78.0 & 21.0 & \textbf{6.15} & \textbf{7.0} & \textbf{73.4} & 20.6 \\
$-$ Pivot ($\gamma{=}0$) & 6.81 & 25.8 & 67.2 & 16.5 & 5.85 & 13.5 & \textbf{79.9} & 23.0 & 5.54 & 5.9 & 71.4 & \textbf{21.7} \\
$-$ CMDS ($\beta{=}0$) & 6.09 & 27.3 & 51.1 & 21.8 & 5.10 & 17.5 & 78.5 & \textbf{24.4} & 5.79 & 6.3 & 71.6 & 17.2 \\
\bottomrule
\end{tabular}
\end{table}
 
\textbf{(1) Pivot attribution drives answer precision on grounded reasoning.}
We removed the pivot signal ($\gamma_{\text{dyn}}{=}0$) that drops LLM-J by $0.62$ points on Physical (7.43$\to$6.81) and $6.02$ on Commonsense (6.02$\to$5.85), with $EM_{a}$ falling by $32\%$ points on Physical (38.0$\to$25.8) and $41\%$ on Quantitative (23.0$\to$13.5). The mechanism's impact scales with how tightly reasoning steps depend on visual grounding: Physical reasoning requires the model to chain observations (``I see a white shirt'') into conclusions (``Therefore the lead player wears white''), and pivot tokens at these transitions carry the cross-modal signal. Commonsense reasoning, which relies somewhat on distributed knowledge, shows the smallest LLM-J drop ($0.61$ points).

\textbf{(2) CMDS protection is essential for cross-modal integrity.}
Removing CMDS ($\beta{=}0$) produces the steepest $SEM_{c+a}$ drop in the entire ablation: Physical crashes from $69.9$ to $51.1$ ($-27\%$), indicating severe disruption of vision-language alignment. LLM-J falls from $7.43$ to $6.09$ ($-1.34$ points) and Formatting Score from $26.5$ to $21.8$. Without CMDS, the pruner distributes cuts uniformly across layers, removing units in the vision-language fusion band (layers $11-18$) that the sliding window ablation (Fig~\ref{fig:cmds_plot}(b)) identifies as the cross-modal bottleneck. On Commonsense, the $SEM_{c+a}$ and LLM-J drops are smaller ($1.8$ and $0.36$ LLM-J points respectively), consistent with these domains relying less  on visual grounding.
 
\textbf{(3) Random pivots confirm that pivot \emph{location} matters, not just pivot \emph{existence}.}
\label{app:random_pivots}
We replaced real pivot tokens with randomly selected positions provides a critical control: if \method{}'s gains came merely from restricting the attribution window to \emph{any} token subset, random pivots should perform comparably. Table~\ref{tab:random_pivot_experiment} shows they do not.
 
\begin{table}[h]
\centering
\caption{Random vs.\ real pivot selection on Qwen2.5-VL-7B at $50\%$ pruning. Random pivots recover only partial gains over no-pivot, confirming that correct identification of reasoning transitions drives \method{}'s advantage where \textit{J} stands for LLM-Judge score, \textit{$EM_{a}$} for string exact match of final Answer, \textit{$SEM_{c+a}$} for Semantic Similarity (cosine) of the complete \textit{CoT + final answer}, \textit{Fmt} for Formatting Score which measures if the model can generate step-by-step formatted reasoning steps}
\label{tab:random_pivot_experiment}
\small
\setlength{\tabcolsep}{3pt}
\begin{tabular}{l rrrr rrrr rrrr}
\toprule
& \multicolumn{4}{c}{\textbf{Physical}} & \multicolumn{4}{c}{\textbf{Quantitative}} & \multicolumn{4}{c}{\textbf{Commonsense}} \\
\cmidrule(lr){2-5} \cmidrule(lr){6-9} \cmidrule(lr){10-13}
\textbf{Pivot Selection} & J & $EM_{a}$ & $SEM_{c+a}$ & Fmt & J & $EM_{a}$ & $SEM_{c+a}$ & Fmt & J & $EM_{a}$ & $SEM_{c+a}$ & Fmt \\
\midrule
No Pivot ($\gamma{=}0$) & 7.13 & 25.8 & 67.2 & 16.5 & 6.02 & 13.5 & \textbf{79.9} & \textbf{23.0} & 5.54 & 5.9 & 71.4 & \textbf{21.7} \\
Random Pivot & 6.95 & 29.2 & \textbf{73.1} & 17.8 & 5.79 & 11.9 & 78.6 & 22.5 & 5.53 & 6.0 & 72.0 & 20.8 \\
Real Pivot (default) & \textbf{7.85} & \textbf{38.0} & 69.9 & \textbf{26.5} & \textbf{6.36} & \textbf{23.0} & 78.0 & 21.0 & \textbf{6.15} & \textbf{7.0} & \textbf{73.4} & 20.6 \\
\bottomrule
\end{tabular}
\end{table}
 
Random pivots recover $EM_{a}$ only partially over the no-pivot baseline on Physical ($29.2$ vs $25.8$, compared to $38.0$ for real pivots) and actually \emph{hurt} EM on Quantitative ($11.9$ vs $13.5$ for no-pivot). LLM-J follows the same pattern: random pivots reach $6.95$ on Physical (vs $7.85$ for real, $71.3$ for none) but only $5.53$ on Commonsense (vs $6.15$ for real, $5.54$ for none), performing no better than removing pivots entirely.
The gap between random and real pivots, particularly the $8.8$-point $EM_{a}$ gap on Physical and $11.1$-point gap on Quantitative confirms that \method{}'s gains come from correctly identifying reasoning-transition tokens, not from the mere act of restricting the attribution window to a subset of positions.
Interestingly, random pivots achieve the highest Semantic Similarity on Physical ($73.1$, above both real pivots at $69.9$ and no-pivot at $67.2$), suggesting they preserve broad semantic content while failing to protect the specific inter-step transitions that $EM_{a}$ and LLM-J measure.

\subsection{Pivot Marker Taxonomy for Transition Token Detection}
\label{app:pivot_markers}
 
We detect reasoning transition tokens by scanning the generated CoT sequence for surface-level markers that signal a shift in the model's reasoning state. These markers fall into two broad classes\emph{structural delimiters} and \emph{logical connectives} each targeting a different surface realisation of the same underlying event: the model completing one reasoning step and initiating the next. Structural delimiters are explicit formatting cues that CoT prompts typically encourage, such as numbered prefixes and terminal answer markers. Logical connectives are natural-language words and phrases that signal causal or conclusory transitions; they capture the same step boundaries in models whose outputs are less rigidly formatted. For each detected marker, we construct a transition window $\mathcal{M}_W = \{t : |t - i| \leq W\}$ of half-width $W{=}8$ tokens centered on the marker position $i$, and restrict the pivot attribution loss to token positions within this window (Eq.~8 in the main paper). When no marker from either class is found in a sample, approximately $15\%$ of cases, we fall back to equal-interval partitioning of the response into three segments and treat the segment boundaries as pivot positions. Table~\ref{tab:pivot_markers} provides the complete taxonomy of markers used in our implementation, together with representative examples drawn from generated CoT traces. The taxonomy is intentionally broad so that it generalizes to CoT formats produced by different models and prompting conventions; adding or removing individual patterns has negligible effect on results, as the random-pivot ablation in App.~\ref{tab:random_pivot_experiment} confirms.

\begin{table}[h]
\centering
\caption{Taxonomy of pivot markers used for reasoning transition token detection. Markers in both classes are matched case-insensitively via regular expressions. The window $\mathcal{M}_W$ of half-width $W{=}8$ is constructed around each detected position.}
\label{tab:pivot_markers}
\small
\setlength{\tabcolsep}{4pt}
\renewcommand{\arraystretch}{1.1}

\begin{tabularx}{\columnwidth}{
l 
l 
>{\raggedright\arraybackslash\ttfamily}X 
>{\raggedright\arraybackslash}X
}
\toprule
\textbf{Class} & \textbf{Sub-type} & \textbf{Patterns (regex)} & \textbf{Example in CoT trace} \\
\midrule

\multirow{10}{*}{\parbox{2.1cm}{\textbf{Structural\\Delimiters}}}
 & Numbered step & \^{}(\textbackslash d+)[.)]\textbackslash s & \texttt{1. I observe a red ball...} \\
\cmidrule{2-4}
 & Labelled step & Step\textbackslash s*\textbackslash d+\textbackslash s*[:\textendash] & \texttt{Step 2: The object is circular...} \\
\cmidrule{2-4}
 & Sub-step & (\textbackslash d+)\.(\textbackslash d+)\textbackslash s & \texttt{1.1 Examining the left panel...} \\
\cmidrule{2-4}
 & Final answer & Final Answer[:\textendash] & \texttt{Final Answer: three} \\
\cmidrule{2-4}
 & Answer marker & \textbackslash bAnswer\textbackslash b\textbackslash s*[:\textendash] & \texttt{Answer: blue} \\
\cmidrule{2-4}
 & Conclusion label & (Conclusion|Summary)[:\textendash] & \texttt{Conclusion: the taller object...} \\
\cmidrule{2-4}
 & Think/reason tag & <think>|</think> & \texttt{</think> The answer is four.} \\

\midrule

\multirow{12}{*}{\parbox{2.1cm}{\textbf{Logical\\Connectives}}}
 & Causal conclusion & \textbackslash b(Therefore|Thus|Hence)\textbackslash b & \texttt{Therefore, the count is two.} \\
\cmidrule{2-4}
 & Consequence & \textbackslash b(So|Consequently|As a result)\textbackslash b & \texttt{So the glass will spill over.} \\
\cmidrule{2-4}
 & Inference & \textbackslash b(This means|This implies|This suggests)\textbackslash b & \texttt{This means the car is moving left.} \\
\cmidrule{2-4}
 & Deduction & \textbackslash b(We can (therefore|thus) conclude)\textbackslash b & \texttt{We can therefore conclude that...} \\
\cmidrule{2-4}
 & Summary & \textbackslash b(In (summary|conclusion)|To summarize)\textbackslash b & \texttt{In summary, three forks are visible.} \\
\cmidrule{2-4}
 & Transition adverb & \textbackslash b(Next|Now|Moving on|Finally)\textbackslash b & \texttt{Next, I identify the spatial layout.} \\
\cmidrule{2-4}

& Contrastive pivot &
\makecell[l]{\textbackslash b(However|Nevertheless|\\Despite this)\textbackslash b}
& \texttt{However, the shadow indicates depth.} \\
\cmidrule{2-4}

& Additive reasoning &
\makecell[l]{\textbackslash b(Furthermore|\\Additionally|Moreover)\textbackslash b}
& \texttt{Furthermore, the label reads ``exit''.} \\

\midrule
\multicolumn{2}{l}{\textbf{Fallback (no marker found)}} 
& Equal-interval partition 
& Response split into 3 equal segments; boundaries treated as pivots. \\

\bottomrule
\end{tabularx}
\end{table}

\subsection{Global allocation is indispensable.}
We replaced global knapsack allocation with uniform layerwise pruning, where every layer receives the same pruning ratio, causes the most catastrophic failure across all ablations (Table~\ref{tab:global_vs_layer}).
 
\begin{table}[h]
\centering
\caption{Global vs.\ layerwise allocation on Qwen2.5-VL-7B at $50\%$ pruning. Layerwise allocation collapses reasoning across all domains despite retaining the same total parameter count, where \textit{J} stands for LLM-Judge score, \textit{$EM_{a}$} for string exact match of final Answer, \textit{$EM_{c+a}$} for Semantic Similarity (cosine) of the complete \textit{CoT + final answer}, \textit{Fmt} for Formatting Score which measures if the model can generate step-by-step formatted reasoning steps}
\label{tab:global_vs_layer}
\small
\setlength{\tabcolsep}{3pt}
\begin{tabular}{l rrrr rrrr rrrr}
\toprule
& \multicolumn{4}{c}{\textbf{Physical}} & \multicolumn{4}{c}{\textbf{Quantitative}} & \multicolumn{4}{c}{\textbf{Commonsense}} \\
\cmidrule(lr){2-5} \cmidrule(lr){6-9} \cmidrule(lr){10-13}
\textbf{Allocation} & J & $EM_{a}$ & $SEM_{c+a}$ & Fmt & J & $EM_{a}$ & $SEM_{c+a}$ & Fmt & J & $EM_{a}$ & $SEM_{c+a}$ & Fmt \\
\midrule
Global (default) & \textbf{7.43} & \textbf{38.0} & \textbf{69.9} & \textbf{26.5} & \textbf{6.02} & \textbf{23.0} & \textbf{78.0} & \textbf{21.0} & \textbf{6.15} & \textbf{7.0} & \textbf{73.4} & \textbf{20.6} \\
Layerwise & 4.40 & 3.03 & 66.2 & 1.5 & 3.89 & 2.2 & 62.1 & 4.1 & 3.85 & 3.01 & 57.3 & 10.8 \\
\bottomrule
\end{tabular}
\end{table}
 
LLM-J collapses from $7.43$ to $4.40$ on Physical ($-3.30$ points), from $6.02$ to $3.85$ on Quantitative ($-2.13$), and from $6.15$ to $3.85$ on Commonsense ($-2.30$). EM drops to the lowest across all domains ($3.03$, $2.2$, $3.01$) and Formatting Score falls to $\sim1.5$, indicating that the pruned model can barely produce coherent CoT structure.
Layerwise allocation forces equal pruning ratios on every layer, which removes a disproportionate number of cheap MLP neurons in some layers while under-pruning expensive GQA groups in others, a particularly destructive mismatch given the $\sim$250$\times$ per-unit cost disparity in Qwen2.5-VL-$7B$ ($5.2M$ parameters per GQA group vs $21K$ per MLP neuron), $\sim$85$\times$ in Gemma $3$ $4B$ ($0.66M$ vs $7.7K$), and $\sim$21$\times$ in Qwen2-VL-2B ($98K$ vs $4.6K$). This ablation delivered us the strongest reason for the global knapsack formulation: even with pivot attribution and CMDS protection intact, layerwise allocation renders the model non-functional. The knapsack solver allows the pruner to concentrate its budget on high-value, high-CMDS layers while aggressively compressing low-importance regions, a flexibility that uniform allocation fundamentally cannot provide.

\subsection{Transition window size exhibits a clear optimum at $W{=}8$.}
Table~\ref{tab:window} compares three window sizes. Both $W{=}4$ and $W{=}16$ degrade performance relative to $W{=}8$, but through different mechanisms.
 
\begin{table}[H]
\centering
\caption{Effect of transition window size $W$ on Qwen2.5-VL-7B at 50\% pruning.}
\label{tab:window}
\small
\setlength{\tabcolsep}{3pt}
\begin{tabular}{l rrrr rrrr rrrr}
\toprule
& \multicolumn{4}{c}{\textbf{Physical}} & \multicolumn{4}{c}{\textbf{Quantitative}} & \multicolumn{4}{c}{\textbf{Commonsense}} \\
\cmidrule(lr){2-5} \cmidrule(lr){6-9} \cmidrule(lr){10-13}
$W$ & J & $EM_{a}$ & $SEM_{c+a}$ & Fmt & J & $EM_{a}$ & $SEM_{c+a}$ & Fmt & J & $EM_{a}$ &  $SEM_{c+a}$ & Fmt \\
\midrule
$W{=}4$ & 7.70 & 27.4 & \textbf{75.7} & 19.9 & 6.0 & 11.4 & \textbf{78.1} & \textbf{31.8} & 5.84 & 6.6 & \textbf{73.6} & \textbf{21.9} \\
$W{=}8$ (default) & 7.85 & \textbf{38.0} & 69.9 & \textbf{26.5} & \textbf{6.36} & \textbf{23.0} & 78.0 & 21.0 & \textbf{6.15} & \textbf{7.0} & 73.4 & 20.6 \\
$W{=}16$ & \textbf{7.72} & 30.5 & 67.2 & 23.9 & 5.60 & 6.7 & 77.2 & 21.9 & 5.59 & 7.0 & 71.4 & 20.6 \\
\bottomrule
\end{tabular}
\end{table}
\vspace{-0.7em}
 
$W{=}4$ captures too few context tokens around each pivot, reducing the signal-to-noise ratio of the attribution score: $EM_{a}$ drops by $11\%$ on Physical and $12\%$ on Quantitative relative to $W{=}8$.
$W{=}16$ dilutes the pivot signal by including too many non-transition tokens, collapsing toward global attribution: Quantitative $EM_{a}$ drops from $23.0$ to $6.7$ ($-16.3$) and LLM-J from $6.36$ to $5.60$, while Commonsense LLM-J falls from $6.15$ to $3.85$. An instructive counter-pattern appears in Semantic Similarity, which \emph{peaks} at $W{=}4$ on Physical ($75.7$ vs $69.9$ for $W{=}8$): the narrower window preserves overall semantic content, responses remain topically relevant, but sacrifices the precise inter-step transitions that $EM_{a}$ and LLM-J measure. This dissociation reinforces that embedding similarity alone is insufficient for evaluating CoT preservation. The default $W{=}8$ achieves the best $EM_{a}$ and LLM-J across all three domains, striking the optimal balance between capturing sufficient transition context and maintaining pivot specificity.

\subsection{Output Distribution Fidelity: Logit Divergence Analysis}
\label{app:logit_divergence}
Standard evaluation metrics such as LLM-Judge, Exact Match, and Semantic Similarity operate on \emph{generated text} and therefore capture downstream effects of pruning only after auto-regressive decoding amplifies or dampens initial distortions. To measure how faithfully each pruning method preserves the dense model's predictive behaviour at the source, we compute per-token KL divergence directly on the output logit distributions across all three reasoning
domains.

\begin{figure}[t]
\centering
\includegraphics[width=\linewidth]{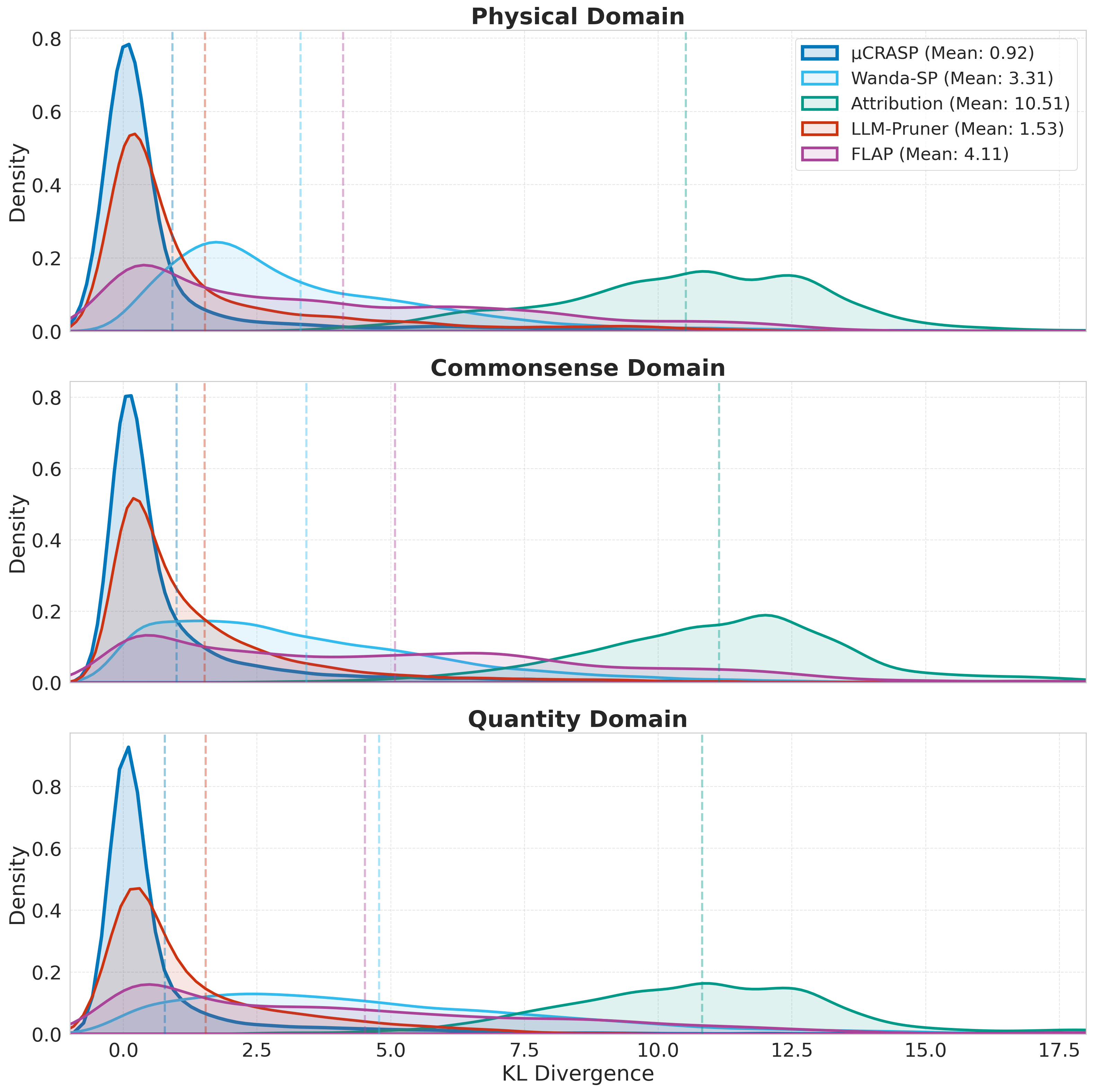}
\caption{Per-token KL divergence distributions
($D_{\mathrm{KL}}(\text{dense} \| \text{pruned})$) on Qwen2.5-VL-7B at $45\%$ pruning across all three reasoning domains. Dashed vertical lines mark per-method means.\method{} concentrates KL mass near zero (mean $0.92$ nats) in every domain, while baselines exhibit broader and right-shifted distributions. The near-identical distributional structure across Physical, Commonsense, and Quantity confirms that \method{}'s output fidelity is domain-agnostic.}
\label{fig:kl_div}
\end{figure}

\paragraph{Setup.}
We feed $512$ image--question--CoT samples from each domain's calibration set through both the dense (unpruned) Qwen2.5-VL-7B and each pruned variant at $30\%$ ratio. At every answer-token position (i.e., positions where the training label is not masked), we extract the full-vocabulary logit vector, convert it to a probability distribution via softmax, and compute $D_{\mathrm{KL}}(p_{\text{dense}} \| p_{\text{pruned}})$. We restrict the computation to answer tokens because these carry the reasoning and final-answer content; prompt and image tokens are excluded. Token positions that produce NaN or infinite KL values due to numerical edge cases in softmax are filtered out; fewer than $0.1\%$ of positions are affected. The resulting per-token KL distributions are shown in Fig~\ref{fig:kl_div} for all three domains.

\paragraph{Results.}
Table~\ref{tab:kl_summary} reports mean KL divergence per method on the Physical domain; the Commonsense and Quantity panels in Fig~\ref{fig:kl_div} show near-identical distributional structure.
\method{} achieves a mean of $0.92$ nats per token on Physical reasoning, representing a $1.66{\times}$ reduction over LLM-Pruner ($1.53$), a $3.6{\times}$ reduction over Wanda-SP ($3.31$), a $4.5{\times}$ reduction over FLAP ($4.11$), and an $11.4{\times}$ reduction over Attribution Pruning ($10.51$).

\begin{table}[h]
\centering
\caption{Mean per-token KL divergence ($D_{\mathrm{KL}}(\text{dense} \| \text{pruned})$) on Qwen2.5-VL-7B at $30\%$ pruning, Physical domain. The Commonsense and Quantity domains yield near-identical rankings (Fig~\ref{fig:kl_div}).}
\label{tab:kl_summary}
\small
\begin{tabular}{lcc}
\toprule
\textbf{Method} & \textbf{Mean KL (nats)} & \textbf{Ratio vs.\ \method{}} \\
\midrule
\method{}           & $\mathbf{0.92}$ & $1.0{\times}$   \\
LLM-Pruner          & $1.53$          & $1.7{\times}$   \\
Wanda-SP            & $3.31$          & $3.6{\times}$   \\
FLAP                & $4.11$          & $4.5{\times}$   \\
Attribution Pruning & $10.51$         & $11.4{\times}$  \\
\bottomrule
\end{tabular}
\end{table}

\paragraph{Domain-consistency of distributional fidelity.}
The most important observation from Fig~\ref{fig:kl_div} is that the relative ordering and distributional shapes are essentially invariant across all three domains. In every panel, \method{}'s KL density concentrates sharply near zero (peak density ${\sim}0.80$), while Attribution Pruning produces a broad bimodal distribution centered near $10$--$11$ nats with substantial mass extending past $15$ nats. LLM-Pruner consistently sits closest to \method{} but with a noticeably heavier right tail, Wanda-SP and FLAP occupy intermediate positions, and their cross-domain ordering is preserved. This domain-invariance directly parallels the domain-agnostic LLM-J gains reported in Section~\ref{sec:results}: the same architectural structures that \method{} protects via CMDS and pivot attribution are the ones responsible for logit-level fidelity regardless of reasoning type.

\paragraph{Connection to reasoning preservation.}
In auto-regressive generation, each token's probability distribution conditions all subsequent predictions. When pruning distorts the logit distribution at a given position, the error compounds through the remaining sequence: the model selects a different token, which shifts the context for the next prediction, producing a cascading deviation from the dense model's reasoning trajectory. \method{}'s low and tightly concentrated KL divergence ensures that the pruned model's token-by-token decisions remain closely aligned with the dense model's across all three domains, preventing the cascading errors that cause baselines
to produce logically incoherent reasoning chains despite generating fluent text.
This observation complements the pivot attribution analysis (Section~\ref{sec:analysis}): pivot tokens at reasoning-step boundaries are precisely the positions where even small logit distortions redirect the entire downstream reasoning trajectory, and \method{}'s concentrated low-KL profile protects these critical positions most effectively and does so consistently whether the task demands visual grounding, numerical precision, or world knowledge.

\section{\method{} Algorithm}
\label{app:algorithm}

\begin{algorithm}[H]
\caption{\method{}: Modality-and-Reasoning-Aware Global Knapsack Pruning for VLMs}
\label{alg:vcot_prune}
\small
\begin{algorithmic}[1]
\REQUIRE VLM $\mathcal{M}$, Calibration Set $\mathcal{D}$, Pruning Ratio $S$, $\gamma_{\text{base}}$, $\rho$, Window $W$, Pivot Extractor $\Phi_{\text{pivot}}$
\ENSURE Pruned VLM $\mathcal{M}'$
\STATE \textbf{\% Phase 1: Global Attribution}
\STATE $A_{\text{global}} \leftarrow 0$
\FOR{each batch $b \in \mathcal{D}$}
    \STATE Compute $\mathcal{L}_{\text{LM}}$; backpropagate; \quad $A_{\text{global}} \leftarrow A_{\text{global}} + |\mathcal{W} \odot \nabla_\mathcal{W} \mathcal{L}|$
\ENDFOR
\STATE $\bar{A}_{\text{global}} \leftarrow A_{\text{global}} / |\mathcal{D}|$
\STATE \textbf{\% Phase 2: Trajectory Pivot Attribution}
\STATE $A_{\text{pivot}} \leftarrow 0$, $N_{\text{trans}} \leftarrow 0$
\FOR{each batch $b \in \mathcal{D}$}
    \STATE $\mathcal{I}_{\text{pivot}} \leftarrow \Phi_{\text{pivot}}(b)$
    \IF{$\mathcal{I}_{\text{pivot}} \neq \emptyset$}
        \STATE $\mathcal{M}_W \leftarrow \bigcup_{i \in \mathcal{I}_{\text{pivot}}} \{t : |t-i| \leq W\}$
        \STATE Compute $\mathcal{L}_{\text{pivot}}$ on $\mathcal{M}_W$ (Eq.~\ref{eq:transition_loss}); backpropagate
        \STATE $A_{\text{pivot}} \leftarrow A_{\text{pivot}} + |\mathcal{W} \odot \nabla_\mathcal{W} \mathcal{L}_{\text{pivot}}|$; \quad $N_{\text{trans}} \leftarrow N_{\text{trans}} + 1$
    \ENDIF
\ENDFOR
\STATE $\bar{A}_{\text{pivot}} \leftarrow A_{\text{pivot}} / N_{\text{trans}}$
\STATE \textbf{\% Phase 3: Layer-Level Profiling}
\FOR{each layer $L$}
    \STATE $\text{Sens}(L) \leftarrow \sqrt{\mathbb{E}[\|h_L\|_2^2]}$; \quad $\text{CMDS}(L) \leftarrow$ Eq.~\ref{eq:cmds}
\ENDFOR
\STATE \textbf{\% Phase 4: Fusion \& Protection}
\STATE $\gamma_{\text{dyn}} \leftarrow \gamma_{\text{base}} (1-S)^\rho$; \quad $A_{\text{fused}} \leftarrow (1{-}\gamma_{\text{dyn}})\bar{A}_{\text{global}} + \gamma_{\text{dyn}} \bar{A}_{\text{pivot}}$
\STATE Compute $\mathcal{P}(L)$ for all layers (Eq.~\ref{eq:protection})
\STATE \textbf{\% Phase 5: Knapsack Packing}
\FOR{each unit $u$}
    \STATE $E_u \leftarrow A_{\text{fused}}^{(u)} \times \mathcal{P}(L_u) / c(u)$
\ENDFOR
\STATE Sort by $E_u$; greedily pack into $\mathcal{K}$ until $\sum c(u) \leq (1{-}S) \times \text{TotalParams}$
\STATE \textbf{\% Phase 6: Safety Enforcement}
\STATE Override $\mathcal{K}$ to satisfy min-retention (Eq.~\ref{eq:safety}); \quad $\mathcal{M}' \leftarrow \text{PhysicalPrune}(\mathcal{M}, \mathcal{K})$
\RETURN $\mathcal{M}'$
\end{algorithmic}
\end{algorithm}

\subsection{Global Attribution Scoring}
\label{sec:global_attr}
In autoregressive generation, a response $Y = (y_1, \ldots, y_T)$ is produced
token-by-token. Aggregating the first-order Taylor importance
$\Delta\mathcal{L} \approx \mathbf{w}_u^\top \nabla_{\mathbf{w}_u} \mathcal{L}$
over all token positions yields the global score:
\begin{equation}
\label{eq:global_attr}
I_{\text{global}}(u) = \sum_{t=1}^{T} \left| \mathbf{w}_u^\top \nabla_{\mathbf{w}_u} \ell(y_t \mid y_{<t}) \right|,
\end{equation}
accumulated over the calibration dataset $\mathcal{D}$.

For \textbf{MLP units}, pruning neuron $j$ removes its parameters across gate, up, and down projections. Its importance aggregates across all three:
\begin{equation}
\label{eq:mlp_attr}
I_{\text{global}}^{\text{MLP}}(j) = \sum_{\text{proj} \in \{g,u,d\}} \sum_k \left| W_{\text{proj}}^{jk} \cdot \frac{\partial \mathcal{L}}{\partial W_{\text{proj}}^{jk}} \right|.
\end{equation}

For \textbf{attention units}, let $\mathcal{H}_u$ denote the parameter indices associated with unit $u$ (a single head in MHA, or a positional group in GQA). Its importance is:
\begin{equation}
\label{eq:attn_attr}
I_{\text{global}}^{\text{attn}}(u) = \sum_{d \in \mathcal{H}_u} \left( \left| W_Q^d \cdot \nabla_{W_Q^d} \mathcal{L} \right| + \left| W_O^d \cdot \nabla_{W_O^d} \mathcal{L} \right| \right).
\end{equation}

\subsection{Trajectory Pivot Attribution}
\label{sec:tpa}

Equation~\ref{eq:global_attr} aggregates importance uniformly across all $T$ token positions, which is suboptimal for CoT reasoning. Instead, CoT generation can be viewed as a trajectory over latent reasoning states $Z = (z_1, \ldots, z_K)$, where most tokens describe intermediate states with predictable structure, while a small subset corresponds to transitions between reasoning steps. These transition tokens are sparse (${\sim}5$\% of the sequence) yet disproportionately informative for reasoning progression, and their contribution is diluted under global aggregation.

\textbf{Transition region estimation.}\quad 
We model CoT as a sequence of state transitions:
$
\label{eq:cot_markov}
p(y \mid x) = \prod_{k=1}^{K} p(z_k \mid z_{<k}, x) \cdot p(a \mid z_{1:K}, x).
$
Each conditional $p(z_k \mid z_{<k}, x)$ is governed by localized regions where the model transitions between reasoning states. Rather than assuming a fixed format, we estimate a set of transition indices $\mathcal{I}_{\text{pivot}}$ as a coarse approximation to these boundaries. 

In practice, we identify candidate transitions using lightweight structural cues when available (e.g., step delimiters or discourse shifts), and otherwise fall back to uniform segmentation of the sequence. This procedure is not intended to be exact, but to provide a weak localization of transition regions that is agnostic to specific CoT formats. 

Around each estimated transition, we construct a local window:
\begin{equation}
\label{eq:transition_window}
\mathcal{M}_W = \bigcup_{i \in \mathcal{I}_{\text{pivot}}} \{t : |t - i| \leq W\},
\end{equation}
which captures the tokens most relevant for reasoning state updates. As shown in our ablations, the effectiveness of this approach does not depend on precise boundary identification, but rather on concentrating attribution around transition regions.


\textbf{Pivot attribution score.}\quad We restrict the loss to transition tokens:
\begin{equation}
\label{eq:transition_loss}
\mathcal{L}_{\text{pivot}} = -\sum_{t \in \mathcal{M}_W} \log p_\theta(y_t \mid y_{<t}, x^v, x^t),
\end{equation}
and compute first-order Taylor importance on this restricted loss, normalized by the number of pivots:
\begin{equation}
\label{eq:pivot_attr}
A_{\text{pivot}}(u) = \frac{1}{|\mathcal{I}_{\text{pivot}}|} \sum_{t \in \mathcal{M}_W} \left| \mathbf{w}_u^\top \nabla_{\mathbf{w}_u} \ell(y_t \mid y_{<t}) \right|.
\end{equation}

Trajectory Pivot Attribution can be interpreted as a conditional variant of Taylor-based importance estimation \citep{molchanov2019importance}, where the loss landscape is restricted to transition tokens. This aligns with information-theoretic perspectives on compression: effective pruning should preserve components that retain high mutual information with future reasoning states while discarding those that merely decode low-entropy syntactic patterns \citep{tishby2000information}.

Pivot extraction serves as a proxy for latent reasoning state transitions. While explicit markers make these boundaries observable, the underlying assumption is that reasoning trajectories contain sparse transition points irrespective of surface form. In principle, pivot identification can be replaced with more general boundary detection methods (e.g., entropy peaks, attention shifts, or learned segmentation), making the approach applicable beyond explicitly structured CoT traces.

\subsection{Cross-Modal Dependency Score (CMDS)}
\label{sec:cmds}

Existing pruning methods do not explicitly account for cross-modal interactions between vision and language tokens. They may remove components critical for aligning visual and textual representations, leading to degraded multimodal reasoning. We seek a principled measure that quantifies how strongly a layer differentiates between modalities.

\textbf{Distributional formulation.}\quad We formulate cross-modal dependency as a divergence between vision and text token activation distributions. Let $\mathbb{P}_V$ and $\mathbb{P}_T$ denote these distributions at layer $L$. We quantify their discrepancy through Maximum Mean Discrepancy (MMD) \citep{gretton2012kernel, gretton2006kernel}, a kernel-based two-sample test:
\begin{equation}
\label{eq:mmd_general}
\text{MMD}[\mathcal{H}, \mathbb{P}_V, \mathbb{P}_T] = \sup_{\|f\|_{\mathcal{H}} \leq 1} \left( \mathbb{E}_{h_v}[f(h_v)] - \mathbb{E}_{h_t}[f(h_t)] \right).
\end{equation}
MMD equals zero if and only if $\mathbb{P}_V = \mathbb{P}_T$ for characteristic kernels \citep{gretton2012kernel}. We instantiate with a linear kernel $\kappa(x,y)=x^\top y$, yielding a tractable estimate:
\begin{equation}
\label{eq:mmd_linear}
\text{MMD}_L = \left\| \mathbb{E}[h_v] - \mathbb{E}[h_t] \right\|_2.
\end{equation}

To obtain a scale-invariant measure, we normalize by the expected activation magnitude:
\begin{equation}
\label{eq:cmds}
\text{CMDS}(L) = \frac{\left\| \mathbb{E}[\text{Act}_{\text{vis}}^{(L)}] - \mathbb{E}[\text{Act}_{\text{txt}}^{(L)}] \right\|_2}{\mathbb{E}\left[\|\text{Act}_{\text{all}}^{(L)}\|_2\right] + \epsilon}.
\end{equation}
High CMDS values indicate strong modality-specific transformations (critical cross-modal integration), while low values correspond to modality-agnostic processing. Empirically, CMDS peaks in middle layers across all four VLMs we evaluate (Fig.\ref{fig:cmds_plot}).

\subsection{Layer Output Sensitivity}
\label{sec:sensitivity}

Second-order pruning methods estimate importance using curvature information (e.g., inverse Hessian) \citep{hassibi1993optimal, frantar2023sparsegpt}, but are computationally infeasible for large VLMs. We adopt a lightweight proxy based on forward perturbation in the Transformer residual stream.

In a Transformer, each layer contributes additively: $x_{\ell+1} = x_\ell + f_\ell(x_\ell)$. Pruning units in layer $L$ perturbs its output $h_L$, introducing an error $\delta_L$ that propagates through subsequent layers. The magnitude of $\delta_L$ scales with $\|h_L\|_2$, implying that layers producing larger activations are more sensitive to pruning. Moreover, LayerNorm does not fully absorb such perturbations: it normalizes the full vector, so a large $\delta_L$ changes the \emph{relative} feature magnitudes even after normalization. We define layer sensitivity as:
\begin{equation}
\label{eq:sensitivity}
\text{Sens}(L) = \sqrt{\mathbb{E}_{x \sim \mathcal{D}}\left[\|h_L(x)\|_2^2\right]},
\end{equation}
where $h_L(x)$ is the output of layer $L$'s output projection (\texttt{down\_proj} for MLP, \texttt{o\_proj} for attention). $\text{Sens}(L)$ serves as a zeroth-order proxy for the structural load carried by a layer.

\subsection{Dynamic Gamma Fusion and Global Knapsack}
\label{sec:knapsack}

\paragraph{Pruning Ratio-Conditioned Score Fusion.}

Since Trajectory Pivot Attribution ($A_{\text{pivot}}$) and global attribution ($I_{\text{global}}$) capture complementary aspects of importance, we fuse their scores in a way, that is adaptive to the pruning ratio regime: reasoning structure is more critical at low ratio, while core token prediction dominates under tighter budgets. The fused importance is:
\begin{equation}
\label{eq:fusion}
A_{\text{fused}}^{(u)} = (1 - \gamma_{\text{dyn}}) \cdot \bar{A}_{\text{global}}^{(u)} + \gamma_{\text{dyn}} \cdot \bar{A}_{\text{pivot}}^{(u)},
\end{equation}
where $\bar{A}_{\text{global}}^{(u)}$ and $\bar{A}_{\text{pivot}}^{(u)}$ are normalized importance scores. We model the trade-off via a pruning ratio-conditioned mixing coefficient:
$\gamma_{\text{dyn}}(S) = \gamma_{\text{base}} \cdot (1 - S)^{\rho},$
which smoothly decreases the contribution of the reasoning-specific signal as pruning ratio increases.

\paragraph{Layer-Level Protection Factors.}
To account for cross-layer variation in structural importance, we introduce a multiplicative protection factor:
\begin{equation}
\label{eq:protection}
\mathcal{P}(L) = \big(1 + \alpha \cdot \widetilde{\text{Sens}}(L)\big) \cdot \big(1 + \beta \cdot \widetilde{\text{CMDS}}(L)\big) \cdot \Omega(L),
\end{equation}
where $\widetilde{\text{Sens}}$ and $\widetilde{\text{CMDS}}$ are normalized sensitivity and cross-modal dependency scores.

The structural prior $\Omega(L)$ captures position- and module-specific importance, reflecting empirical observations that certain layers (e.g., early, late, and attention-heavy layers) are more sensitive to pruning. The coefficients $\alpha$ and $\beta$ scale with pruning ratio, enabling stronger differentiation under tighter parameter budgets. Full specification is in Table~\ref{tab:hyperparams_full}.

\subsubsection{Global Knapsack Formulation}

We cast pruning as a global resource allocation problem. Each unit $u$ has value $v(u) = A_{\text{fused}}^{(u)} \cdot \mathcal{P}(L_u)$ and cost $c(u)$, under budget $\mathcal{C}_{\text{target}} = \text{TotalParams} \cdot (1 - S)$. The objective is:
\begin{equation}
\label{eq:knapsack_formal}
\max_{\mathcal{K} \subseteq \bigcup_\ell \mathcal{U}_\ell} \sum_{u \in \mathcal{K}} v(u) \quad \text{subject to} \quad \sum_{u \in \mathcal{K}} c(u) \leq \mathcal{C}_{\text{target}}.
\end{equation}
We solve this via greedy efficiency ranking $E_u = v(u) / c(u)$, packing units until the budget is exhausted. To prevent degenerate solutions, we enforce minimum retention:
\begin{equation}
\label{eq:safety}
|\mathcal{K} \cap \mathcal{U}_L| \geq \eta_L,
\end{equation}
where $\eta_L$ is a pruning ratio-dependent threshold . Structured  pruning removes weight matrix rows/columns, yielding a smaller dense model. The complete algorithm is in Algorithm~\ref{app:algorithm}.


\section{Failure of Magnitude Pruning in VLMs}
\label{sec:appendix_magnitude}
We observe that magnitude-based pruning exhibits catastrophic degradation when applied to VLMs, even at low pruning levels. While the unpruned model maintains reasonable performance (e.g., LLM-J $\approx 69.7$, PPL $\approx 2.21$), pruning rapidly leads to complete collapse of reasoning behavior. At pruning levels as low as $20$--$30\%$, $EM_{a}$ drops to $0$, formatting consistency collapses, and perplexity increases by several orders of magnitude (e.g., $>10^3$ to $>10^5$), indicating loss of coherent generation. 

This behavior is consistent across models and domains, and reflects a fundamental limitation of magnitude pruning: it is agnostic to both task structure and cross-modal dependencies, leading to removal of units critical for maintaining reasoning coherence. Due to this extreme scale mismatch and degenerate outputs, magnitude pruning results are not directly comparable to other methods in the main plots and are therefore reported separately here for completeness.

\begin{table}[h]
\centering
\small
\caption{Representative results of magnitude pruning on Qwen2.5-VL-3B (Commonsense). The method exhibits rapid collapse with increasing pruning level.}
\begin{tabular}{lcccc}
\toprule
\textbf{Levels} & \textbf{$EM_{a}$} & \textbf{$EM_{c+a}$} & \textbf{PPL} & \textbf{Formatting} \\
\midrule
0\%   & 39.65 & 81.98 & 2.21 & 99.87 \\
20\%  & 0.00  & 14.28 & 1.97$\times10^3$ & 0.00 \\
30\%  & 0.00  & 10.00 & 1.46$\times10^5$ & 0.00 \\
40\%  & 0.00  & 11.30 & 4.41$\times10^5$ & 0.00 \\
\bottomrule
\end{tabular}
\label{tab:magnitude_failure}
\end{table}

\section{Datasets}
\label{app:datasets}
We evaluate \method{} across four benchmarks spanning distinct reasoning domains. Each dataset was selected to test a different facet of multimodal reasoning that pruning must preserve: commonsense inference, quantitative computation, physical attribute recognition.

\paragraph{A-OKVQA (Commonsense Reasoning).}
A-OKVQA \citep{schwenk2022aokvqa} is a crowdsourced benchmark of approximately 25K questions over COCO images that require commonsense and world knowledge to answer. Unlike earlier knowledge-based VQA datasets such as OK-VQA \citep{marino2019okvqa}, A-OKVQA questions generally cannot be answered by querying a knowledge base and instead require multi-step reasoning grounded in the visual scene. We use the official train/validation split and evaluate on the validation set. This dataset serves as our primary benchmark for commonsense reasoning, as the questions naturally elicit multi-step CoT responses.

\paragraph{TDIUC-Quantitative (Quantitative Reasoning).}
To evaluate quantitative reasoning (counting, numerical comparison, and arithmetic inference), we construct a targeted subset from the TDIUC dataset \citep{kafle2017tdiuc}, which organizes VQA into 12 semantically distinct task categories to enable fine-grained analysis of reasoning capabilities. TDIUC explicitly includes counting and numerical reasoning tasks, making it well-suited for our evaluation.

We extract quantitative questions via a multi-stage filtering pipeline designed to balance precision and scalability. First, we apply keyword-based filtering (e.g., “how many”, “how much”, “number of”) to obtain a high-recall candidate pool. Second, we employ a strong LLM-based verifier GPT-4o to classify whether each question requires explicit numerical reasoning, filtering out cases solvable via simple object recognition or memorization.

To assess annotation quality, we perform human validation on a randomly sampled subset (20\%) of the filtered data. The human annotator verifies whether the retained questions genuinely require quantitative reasoning, enabling estimation of the classifier precision and correction of systematic errors. 

\paragraph{TDIUC-Physical (Physical Reasoning).}
To evaluate physical attribute reasoning (e.g., color, shape, size, material, and spatial configuration), we construct a dedicated subset from the TDIUC dataset \citep{kafle2017tdiuc}, which provides fine-grained task categorization for visual question answering.

We extract physical reasoning questions using a multi-stage filtering pipeline designed to balance coverage and precision. First, we apply keyword-based filtering (e.g., ``what color'', ``what shape'', ``how big'', ``what material'') to obtain a high-recall candidate pool. Second, we employ a strong LLM-based verifier (GPT-4o) to classify whether each question genuinely requires reasoning about physical properties, filtering out cases solvable via simple label retrieval or direct object recognition. GPT-4o is a high-capacity multimodal model capable of reasoning across visual and textual inputs, making it well-suited for this verification step.

To assess annotation quality, we perform human validation on a randomly sampled subset (20\%) of the filtered data. Annotators verify whether each question requires reasoning about physical attributes under a consistent labeling protocol. This allows us to estimate the precision of the automated pipeline and identify systematic errors. 

Both of this semi-automated pipeline follows common practice in large-scale dataset curation, combining heuristic filtering with model-based verification and human auditing to ensure both scalability and quality. While keyword filtering may introduce noise and LLM-based verification may inherit model biases, the human-in-the-loop evaluation provides a quantitative estimate of reliability, and our analysis shows that residual noise does not affect the comparative evaluation across pruning methods.

\paragraph{Dataset statistics.}
Table~\ref{tab:dataset_stats} summarizes the key statistics of all four benchmarks.

\begin{table}[h]
\caption{Dataset statistics}
\label{tab:dataset_stats}
\centering
\small
\begin{tabular}{@{}lccccc@{}}
\toprule
\textbf{Dataset} & \textbf{Domain} & \textbf{Train} & \textbf{Val} & \textbf{Test} & \textbf{Source} \\
\midrule
A-OKVQA & Commonsense & 17.1K & 1.1K & 6.7K & COCO \\
TDIUC-Quant. & Quantitative & 15K & 2K & 3.56K & TDIUC \\
TDIUC-Physical & Physical & 15K & 2K & 3K & TDIUC \\
\bottomrule
\end{tabular}
\end{table}

\section{Baselines}
\label{app:baselines}

We compare \method{} against four  baselines that span the major paradigms in structured pruning. All baselines are adapted to perform \emph{structural} pruning (removing entire neurons, attention heads, or GQA groups) to ensure a fair comparison with our method, which also produces dense sub-networks. We focus exclusively on structural pruning baselines, as unstructured methods (e.g., SparseGPT \citep{frantar2023sparsegpt}) produce irregular sparsity patterns that do not yield practical inference speedup without specialized hardware, whereas our goal is to produce smaller dense models deployable on commodity hardware.

\paragraph{(1) Magnitude Pruning.}
The simplest structured pruning baseline removes structural units (neurons or attention groups) with the smallest aggregate $\ell_1$ weight norm \citep{han2015deep, zhu2017prune}. For each MLP neuron $j$, we compute $\sum_k |W_{\text{gate}}^{jk}| + |W_{\text{up}}^{jk}| + |W_{\text{down}}^{kj}|$; for each attention unit, we aggregate across Q and O projection slices. Units are ranked globally and the lowest-scoring ones are removed until the target pruning ratio is reached.

\paragraph{(2) Attribution-Based Pruning.}
Following Yang et al.\ \citep{yang2023attribution}, we use an attribution method to determine which neurons are essential for performing the target task. This approach computes task-specific attribution scores using gradient-based saliency and prunes units with low attribution. Unlike the Taylor score baseline, which uses a single global loss, attribution-based pruning explicitly conditions importance on the downstream task distribution, providing a stronger baseline for task-aware compression.

\paragraph{(3) LLM-Pruner.}
LLM-Pruner \citep{llmpruner} discovers coupled structures via dependency graphs and estimates group importance using a combination of first-order and second-order Taylor approximations. It then performs structural pruning by removing entire dependency groups and recovers performance via LoRA fine-tuning. LLM-Pruner now supports GQA architectures \citep{llmpruner_github}, making it directly applicable to our Qwen-family models. We use the official implementation with default hyperparameters and \texttt{param\_second} importance estimation. Since, we are evaluating one-shot pruning performance without any recovery phase, for this baseline we didn't apply the LORA-recovery phase.

\paragraph{(4) FLAP.}
FLAP \citep{an2024flap} introduces a fluctuation-based structured pruning criterion that evaluates whether the output feature map is easily recoverable when a column of weight is removed. It standardizes importance scores to adaptively determine the global compressed model structure, and adds bias terms to compensate for removed neurons. FLAP also provides a structured adaptation of the Wanda criterion, which we use as our Wanda baseline. We use the official implementation with default settings.

\paragraph{(5) Wanda (Structural).}
Following the structured Wanda variant introduced by An et al.\ \citep{an2024flap} (denoted Wanda-sp in their paper), we adapt the Wanda pruning criterion \citep{wanda} from unstructured to structural pruning. For each structural unit, importance is computed as the product of the unit's weight magnitude and the $L_2$ norm of its corresponding input activations, aggregated across all dimensions of the unit. This provides a gradient-free, computationally efficient baseline.

\paragraph{Baseline implementation notes.}
All baselines use the same calibration set of 128 samples, the same target pruning ratio levels, and the same evaluation protocol as \method{}.


\section{Synthetic CoT Generation}
\label{app:cot_generation}

Our method requires CoT-annotated calibration data, i.e., image-question pairs accompanied by multi-step reasoning traces. Since the target VLMs (2B--7B parameters) do not reliably produce high-quality CoT traces at their native scale, we use a larger teacher model to generate synthetic annotations.

\paragraph{Teacher model.}
We use GPT-4o as the CoT generator. We chose this model for three practical reasons. First, it is a strong, readily available vision-language model with reliable instruction-following behavior. Second, it produces consistent and well-structured reasoning traces, which is important for extracting trajectory-level signals. Finally, GPT-4o is architecturally different from the target models used in our experiments (e.g., Qwen, Gemma), which helps reduce potential bias arising from using the same model family for both data generation and evaluation.

\paragraph{Generation protocol.}
For each image-question pair in the calibration set, we prompt the teacher model with the following system instruction:

\begin{tcolorbox}[
    colback=gray!5,
    colframe=black,
    title=Prompt for CoT Generation,
    sharp corners,
    boxrule=0.5pt
]

You are a helpful AI assistant that solves visual questions with clear step-by-step reasoning.

For each example, you will be provided with the following information:
\begin{enumerate}
    \item Image
    \item Question
    \item Answer (if the answer is incorrect, ignore it and generate the correct answer based on the image and question)
\end{enumerate}

Please generate a reasoning chain with 2--3 concise steps where needed, referencing specific visual cues (``I observe...'', ``Because I see...''), then provide the final answer.

\textbf{Format:}
\begin{verbatim}
Step-by-Step Thought:
1. I observe ...
2. I infer ...
3. Therefore ...

deduced_answer: {final_answer}

\end{verbatim}

\textbf{Constraints:}
\begin{itemize}
    \item Since this is a quantitative reasoning question, write numbers in words (e.g., ``one'', ``two'')
\end{itemize}

\end{tcolorbox}

We generate CoT traces using greedy decoding (temperature = $0$) to ensure reproducibility. To ensure annotation quality, all generated CoT traces were manually reviewed by two human annotators, including one of the authors. The review process verified that the reasoning traces were logically consistent, grounded in the visual content, and aligned with the final answer. Traces containing incomplete reasoning, hallucinated visual evidence, or inconsistent conclusions were discarded and regenerated. This additional verification step ensured that the entire dataset including the calibration set contained reliable reasoning trajectories suitable for pruning analysis.

\paragraph{Calibration set construction.}
From the training split of each dataset, we randomly sample 128 image-question pairs and pair them with their corresponding synthetic CoT traces. This calibration set is used for all scoring phases (global attribution, trajectory pivot attribution, CMDS, and sensitivity profiling). The same calibration set is used across all baselines to ensure fair comparison.

\section{Experimental Settings}
\label{app:settings}

\subsection{Hardware}
\label{app:hardware}

All pruning and evaluation has been carried out in NVIDIA A40 46GB GPUs including the scoring phase of \method{} (global attribution, trajectory pivot attribution, CMDS, and sensitivity profiling) are all done using 2 * 46 GB (A40 Chips). And, for full finetuning of the models with CoT Traces we used NVIDIA H100 GPUs. 

All evaluation (inference on validation sets, perplexity computation, metric calculation) and baseline pruning methods are also executed the same GPUs. Table~\ref{tab:hardware} summarizes the hardware allocation.

\begin{table}[h]
\caption{Hardware allocation across experimental stages.}
\label{tab:hardware}
\centering
\small
\begin{tabular}{@{}lll@{}}
\toprule
\textbf{Stage} & \textbf{Hardware} & \textbf{Approx.\ Time} \\
\midrule
Finetuning (7B model) & 2 * H100 80GB & ${\sim}$ 4 Hr\\
Pruning (7B model) & 2 * A40 46GB & ${\sim}$ 45 min \\
Metrics Evaluation & 2 * A40 46GB & ${\sim}$ 30 min \\
\bottomrule
\end{tabular}
\end{table}

\subsection{Architecture Details}
\label{app:arch}

We evaluate a diverse set of recent vision-language models spanning the Qwen2-VL and Gemma $3$ families. All models follow a modular design consisting of a vision encoder, a language decoder, and a lightweight multimodal fusion mechanism. The Qwen-based models employ a ViT-style vision backbone with 3D patch embeddings and an MLP-based patch merger for modality alignment. In contrast, Gemma models utilize a SigLIP-based vision encoder combined with a pooling-based projection module. Across all models, the language component adopts grouped-query attention (GQA) for efficient decoding. Table~\ref{tab:architecture} summarizes the key architectural differences.

\begin{table*}[t]
\centering
\small
\caption{Architecture comparison of evaluated vision-language models. We report the number of vision encoder and language decoder layers, attention type, vision backbone, and multimodal fusion strategy.}
\label{tab:architecture}

\begin{tabular}{lccccc}
\toprule
\textbf{Model} & \textbf{Enc (Vision)} & \textbf{Dec (LM)} & \textbf{Attention} & \textbf{Vision Backbone} & \textbf{Fusion} \\
\midrule
Qwen2-VL 2B        & 32 & 28 & GQA & Qwen2 ViT         & Patch merger \\
Qwen2.5-VL 3B      & 32 & 36 & GQA & Qwen2.5 ViT       & Patch merger \\
Qwen2.5-VL 7B      & 32 & 28 & GQA & Qwen2.5 ViT       & Patch merger \\
Gemma 3 4B         & 27 & 34 & GQA & SigLIP ViT        & AvgPool projector \\

\bottomrule
\end{tabular}
\end{table*}

\subsection{Hyperparameters}
\label{app:hyperparams_detail}

Table~\ref{tab:hyperparams_full} provides the complete hyperparameter specification for \method{}.

\begin{table}[h]
\caption{Complete hyperparameter settings for \method{}.}
\label{tab:hyperparams_full}
\centering
\small
\begin{tabular}{@{}llc@{}}
\toprule
\textbf{Component} & \textbf{Parameter} & \textbf{Value} \\
\midrule
Trajectory Pivot & Window $W$ & 8 \\
 & Min.\ steps for detection & 2 \\
 & Fallback segmentation & Equal thirds \\
\midrule
Dynamic Gamma & $\gamma_{\text{base}}$ & 0.4 \\
 & Decay $\rho$ & 2 \\
\midrule
Layer Protection & Sensitivity coeff.\ $\alpha$ & $0.3 + 1.5S$ \\
 & CMDS coeff.\ $\beta$ & $0.2 + S$ \\
\midrule
Position Protection & Early layers $N_E$ & 4 \\
 & Final layers $N_F$ & 2 \\
 & Early boost & $1 + 0.5(1 - \ell/N_E)$ \\
 & Final boost & 1.3 \\
\midrule
Structure Protection & Attention boost $\omega$ & 1.8 \\
 & Vision boost $\omega$ & 1.2 \\
\midrule
Safety Constraints & Attn min keep & $\max(2, \lfloor n \cdot \max(0.35, 0.70(1{-}S))\rfloor)$ \\
 & MLP min keep & $\max(1, \lfloor n \cdot \max(0.05, 0.25(1{-}S))\rfloor)$ \\
\midrule
Calibration & Samples per dataset & 128 \\
 & Batch size & 2 \\
 & Sequence length & Model-dependent (max context) \\
\bottomrule
\end{tabular}
\end{table}

\subsection{LLM-as-Judge Protocol}
\label{app:llm_judge}
Traditional evaluation metrics such as Exact Match, BLEU, ROUGE-L, or perplexity primarily measure lexical overlap or token prediction quality, but do not directly assess whether a generated chain-of-thought (CoT) trace remains logically coherent after pruning. In practice, a pruned model may still produce fluent text with reasonable semantic similarity while containing internally inconsistent reasoning steps or unsupported conclusions. To evaluate reasoning preservation more directly, we employ an LLM-as-a-Judge (LLM-J) protocol that scores the logical validity of generated CoT traces.\paragraph{Judge model.}We use \textsc{GPT-3.5-Turbo} as the evaluator model. Importantly, the judge model is distinct from both the CoT generator (\textsc{GPT-4o}) and the target VLMs used in our experiments, reducing potential evaluator bias toward any specific model family or architecture.\paragraph{Evaluation procedure.}For each evaluation sample, the judge model receives:(i) the image-question pair,(ii) the reference reasoning trace,(iii) the model-generated CoT trace, and(iv) the final predicted answer.The judge is instructed to evaluate:\begin{itemize}    \item logical coherence of intermediate reasoning steps,    \item consistency across reasoning transitions,    \item grounding of reasoning in the visual context,    \item alignment between the reasoning trace and final answer.\end{itemize}The judge assigns a continuous score on a $0$--$10$ scale, where higher values indicate stronger logical validity and reasoning coherence.\paragraph{Why a continuous $0$--$10$ scale?}We adopt a continuous scoring scheme instead of binary correctness labels to better capture gradual degradation in reasoning quality under pruning. Structured pruning often produces partially coherent reasoning traces that remain semantically meaningful despite containing minor logical inconsistencies or incomplete conclusions. A continuous scale therefore allows finer-grained assessment of intermediate failure modes that are not captured by discrete correctness metrics.\paragraph{Likert-style interpretation.}To improve interpretability, we additionally provide a Likert-style qualitative interpretation of LLM-J score ranges. This mapping serves only as an interpretation aid and does not alter the underlying quantitative evaluation used throughout the paper.\begin{table}[H]\centering\small\caption{Likert-style interpretation of LLM-J scores.}\label{tab:llmj_likert}\begin{tabular}{ccc}\toprule\textbf{LLM-J Range} & \textbf{Likert Level} & \textbf{Interpretation} \\\midrule0 -- 2   & 1 & Incoherent or logically invalid reasoning \\2 -- 4   & 2 & Major reasoning inconsistencies \\4 -- 6   & 3 & Partially coherent reasoning \\6 -- 8   & 4 & Mostly coherent reasoning \\8 -- 10  & 5 & Strong logical coherence and grounding \\\bottomrule\end{tabular}\end{table}. For example, the LLM-J score of $8.87$ achieved by \method{} on Physical reasoning at $30\%$ pruning level corresponds to Likert level $5$, indicating that the generated reasoning traces remain logically coherent and well grounded despite substantial compression.\paragraph{Human verification.}To verify reliability of the evaluation protocol, a subset of generated reasoning traces was manually inspected by two annotators, including one of the authors. We observed strong qualitative agreement between human judgments and LLM-J scores, particularly in distinguishing coherent reasoning from structurally collapsed outputs.

\subsection{Visual Grounding and Hallucination Analysis}
\label{app:grounding_hallucination}

To further evaluate whether compressed VLMs preserve faithful visual reasoning, we perform an additional analysis of visual grounding and hallucination behavior. Beyond logical coherence, an important failure mode under aggressive pruning is the tendency to generate reasoning traces that are either weakly grounded in the image or contain hallucinated visual details unsupported by the input.

\paragraph{Evaluation protocol.}
For each sample, we provide \textsc{GPT-4o} with:
(i) the input image,
(ii) the question,
(iii) the generated CoT trace, with the final predicted answer.
The evaluator is instructed to assess two complementary properties:

\begin{itemize}
    \item \textbf{Grounding Score}: measures how well the generated reasoning trace is supported by observable visual evidence in the image in the scale of $0$--$10$.
    
    \item \textbf{Hallucination Score}: measures the extent to which the reasoning trace introduces unsupported, fabricated, or contradictory visual details also in the scale of $0$--$10$.
\end{itemize}

\method{} generally preserves visually grounded reasoning even under high pruning ratio, whereas competing baselines frequently exhibit semantic drift, unsupported object descriptions, repetitive degeneration, or complete collapse into incoherent text. For example, FLAP and Wanda-SP often generate syntactically corrupted or repetitive outputs, while LLM-Pruner tends to produce fluent but weakly grounded reasoning traces that ignore critical visual evidence. Attribution-based pruning occasionally retains partial reasoning structure, but often introduces hallucinated visual attributes or incomplete conclusions under higher compression. Representative qualitative examples are provided in App.\S~\ref{app:qualitative}. 

\begin{table}[H]
\centering
\small
\caption{Visual grounding and hallucination analysis at $50\%$ pruning on the Physical domain dataset. Scores are assigned by \textsc{GPT-4o} on a $0$--$10$ scale. Higher Grounding is better; lower Hallucination is better.}
\label{tab:grounding_hallucination}
\begin{tabular}{lcc}
\toprule
\textbf{Method} & \textbf{Grounding Score $\uparrow$} & \textbf{Hallucination Score $\downarrow$} \\
\midrule
\textbf{\method{}} & \textbf{8.6} & \textbf{1.8} \\
LLM-Pruner & 4.9 & 5.7 \\
Wanda-SP & 2.8 & 7.9 \\
Attribution Pruning & 5.6 & 4.8 \\
FLAP & 1.9 & 8.6 \\
\bottomrule
\end{tabular}
\end{table}

The results suggest that preserving reasoning coherence alone is insufficient; effective pruning must additionally maintain alignment between generated reasoning and visual evidence. Methods that ignore cross-modal dependencies tend to hallucinate unsupported objects, attributes, or scene interpretations as compression increases, even when the generated text remains superficially fluent.

\begin{tcolorbox}[
    colback=gray!5,
    colframe=black,
    title=Prompt for Visual Grounding and Hallucination Evaluation,
    sharp corners,
    boxrule=0.5pt
]

You are an expert evaluator for vision-language reasoning systems.

You will be given:
\begin{enumerate}
    \item An image
    \item A question about the image
    \item A generated chain-of-thought (CoT) reasoning trace
    \item A final predicted answer
\end{enumerate}

Your task is to evaluate the quality of the reasoning trace along two dimensions:

\begin{itemize}
    \item \textbf{Grounding Score (0--10):}
    Evaluate how well the reasoning is supported by observable visual evidence in the image. High scores indicate that the reasoning correctly references visible objects, attributes, spatial relations, and scene details.

    \item \textbf{Hallucination Score (0--10):}
    Evaluate the extent to which the reasoning introduces fabricated, unsupported, contradictory, or irrelevant visual details. High scores indicate severe hallucination or semantic drift.
\end{itemize}

\textbf{Evaluation Guidelines:}
\begin{itemize}
    \item Penalize reasoning traces that mention objects, colors, entities, or actions not visible in the image.
    \item Penalize repetitive, degenerate, or logically inconsistent reasoning.
    \item Reward concise reasoning grounded in explicit visual observations.
    \item Consider both the reasoning trace and the final answer jointly.
\end{itemize}

\textbf{Output Format:}
\begin{verbatim}
Grounding Score: <0-10>

Hallucination Score: <0-10>

Short Explanation: <brief justification>
\end{verbatim}

\end{tcolorbox}

\section{Dataset Filtering}
\subsection{LLM-based Verification Prompt}
We use a structured prompt to classify whether a question requires quantitative reasoning. The prompt is designed to minimize ambiguity and enforce a binary decision with justification.

\begin{tcolorbox}[
    colback=gray!5,
    colframe=black,
    title=Prompt for Quantitative Reasoning Classification,
    sharp corners,
    boxrule=0.5pt
]
\small

\textbf{Instruction:} You are given a visual question from a VQA dataset. Determine whether answering this question requires \textbf{quantitative reasoning}, such as counting, numerical comparison, or arithmetic inference.

\textbf{Definition of quantitative reasoning:}
A question requires quantitative reasoning if it involves:
\begin{itemize}[nosep]
    \item Counting objects (e.g., ``How many cars are there?'')
    \item Numerical comparison (e.g., ``Are there more cats than dogs?'')
    \item Arithmetic reasoning (e.g., ``How many total items are there if we combine both groups?'')
\end{itemize}

A question does \textbf{not} require quantitative reasoning if it can be answered by:
\begin{itemize}[nosep]
    \item Simple object recognition (e.g., ``What is in the image?'')
    \item Attribute identification (e.g., ``What color is the car?'')
    \item Yes/No questions without counting or numerical comparison
\end{itemize}

\textbf{Question:} \\
\texttt{<INSERT QUESTION HERE>}

\textbf{Output format:}
\begin{itemize}[nosep]
    \item Label: \{Quantitative / Not Quantitative\}
    \item Justification: One sentence explaining the decision
\end{itemize}

\end{tcolorbox}

We retain only questions classified as \texttt{Quantitative}.

\subsection{Human Annotation Protocol}

To validate the quality of the automatically filtered dataset, we conduct human evaluation on the entire dataset.

\paragraph{Annotation task.}
Two Annotators are presented with a question (without model prediction) and asked to determine whether it requires quantitative reasoning according to the same definition used in the LLM prompt.

\paragraph{Guidelines.}
Annotators are instructed to follow these criteria:
\begin{itemize}[nosep]
    \item Label as \textbf{Quantitative} if the answer requires counting, numerical comparison, or arithmetic reasoning.
    \item Label as \textbf{Not Quantitative} if the answer can be obtained via direct perception or recognition without numerical reasoning.
    \item In ambiguous cases, prefer the label corresponding to the \emph{minimal reasoning required}.
\end{itemize}

\paragraph{Quality control.}
To ensure consistency:
\begin{itemize}[nosep]
    \item A subset of samples is independently annotated twice to measure inter-annotator agreement.
    \item Disagreements are resolved through adjudication.
    \item We report Cohen’s $\kappa$ and raw agreement in Appendix~X.
\end{itemize}

\paragraph{Error estimation.}
The human annotations are used to estimate:
\begin{itemize}[nosep]
    \item Precision of the LLM classifier
    \item Error rate of the automated filtering pipeline
\end{itemize}

Following standard human-in-the-loop annotation practices, this hybrid pipeline combines scalable model-based pre-filtering with human validation to ensure high-quality labels while maintaining efficiency.

\section{Discussion}

\textbf{Why is VLM Pruning Harder Than LLM Pruning?}\quad
Indirect evidence from our cross-model results supports the claim that VLM pruning is harder than LLM pruning.
On Gemma-3-4B, all baselines designed for LLMs collapse at 20-25\% pruning ratio on Physical and Commonsense reasoning, despite prior work showing that 4B-parameter text-only models can sustain 30--40\% unstructured pruning ratio with minimal degradation~\citep{frantar2023sparsegpt}.
The key difference is cross-modal coupling: pruning a unit that is unimportant for text generation may be critical for routing information from the vision encoder.
The catastrophic failure of LLM-Pruner on Gemma-3-4B, a method with strong reported results on text-only LLMs, is the most direct evidence for this claim.

\textbf{Perplexity is a misleading proxy for reasoning quality.}\quad
While most of the existing works \citep{ecoflap, llmpruner, shortenedllamadepthpruning} on pruning (Both Structured and Unstructured) benchmarked the pruned models on perplexity, we found that reasoning has little to no correlation with perplexity. For instance on Qwen2.5-VL-7B Physical at 30\%, Base Taylor achieves PPL 1.59 - \emph{lower} than \method{}'s 1.74 - yet its LLM-J is 13.33 versus 88.65. The pruned model generates fluent, low-surprise text that is logically incoherent that is correct tokens without correct reasoning. Conversely, when reasoning genuinely collapses (e.g., LLM-Pruner on Gemma-3-4B at 25\%, LLM-J dropping from 67.71 to 5.35), the perplexity explosion to $>10^6$ occurs only at much higher pruning ratio, long after the model is non-functional. This disconnect, observed across all model sizes, highlights that token-level metrics are insufficient for evaluating CoT preservation and motivates towards LLM-as-Judge protocol.

\section{Limitations.}
(1)~Pivot extraction relies on heuristic pattern matching for step boundary detection; learned segmentation methods could improve robustness for free-form CoT responses.
(2)~The protection factor hyperparameters ($\alpha$, $\beta$, $N_E$, $N_F$, attention boost) are manually chosen, though our ablations (Table~4) demonstrate robustness to individual component removal.
(3)~Evaluation is conducted exclusively on English-language benchmarks; multilingual CoT preservation remains unexplored.
(4)~LLM-as-Judge, while more informative than surface-level metrics like BLEU and F1, may itself have systematic biases that do not capture all reasoning failure modes.
(5)~Our experiments cover VLMs up to 7B parameters; scaling behavior on larger models (e.g., 72B) remains an open question, though prior work on LLM pruning \citep{frantar2023sparsegpt} suggests that larger models are generally more amenable to compression.

\section{Societal Impact}
Our work develops a structured pruning framework that compresses vision-language models while preserving chain-of-thought reasoning quality. The primary positive societal impact is enabling deployment of capable multimodal reasoning models on resource-constrained hardware, reducing the energy cost and carbon footprint of inference and broadening access to VLM technology beyond well-resourced institutions. On the negative side, more efficient VLMs are easier to deploy at scale, which lowers the barrier to misuse for generating disinformation, synthetic media, or automated manipulation of visual content. We do not introduce any new capability beyond what the dense model already possesses; our method only reduces the computational cost of accessing those capabilities. We do not anticipate disproportionate fairness harms specific to our pruning technique, though we acknowledge that compressed models may amplify existing biases present in the base VLM if pruning disproportionately removes parameters responsible for underrepresented concepts. We encourage practitioners deploying pruned VLMs to conduct bias audits appropriate to their application domain.


\section{Model Compression--Performance Curves}
\label{sec:appendix_full_curves}

We provide full compression-performance curves for all evaluated VLMs across three reasoning domains (Physical, Quantitative, Commonsense) and multiple evaluation metrics. Each figure reports performance as a function of pruning ratio, comparing \method{} with all baselines. 

These plots highlight the qualitative differences between methods: while existing pruning approaches exhibit abrupt degradation or collapse beyond moderate pruning level, \method{} maintains stable reasoning performance and degrades more gracefully across metrics. For clarity, each model is shown on a separate page.

\clearpage
\begin{figure*}[t]
    \centering
    \includegraphics[width=\textwidth, height=0.82\textheight, keepaspectratio]{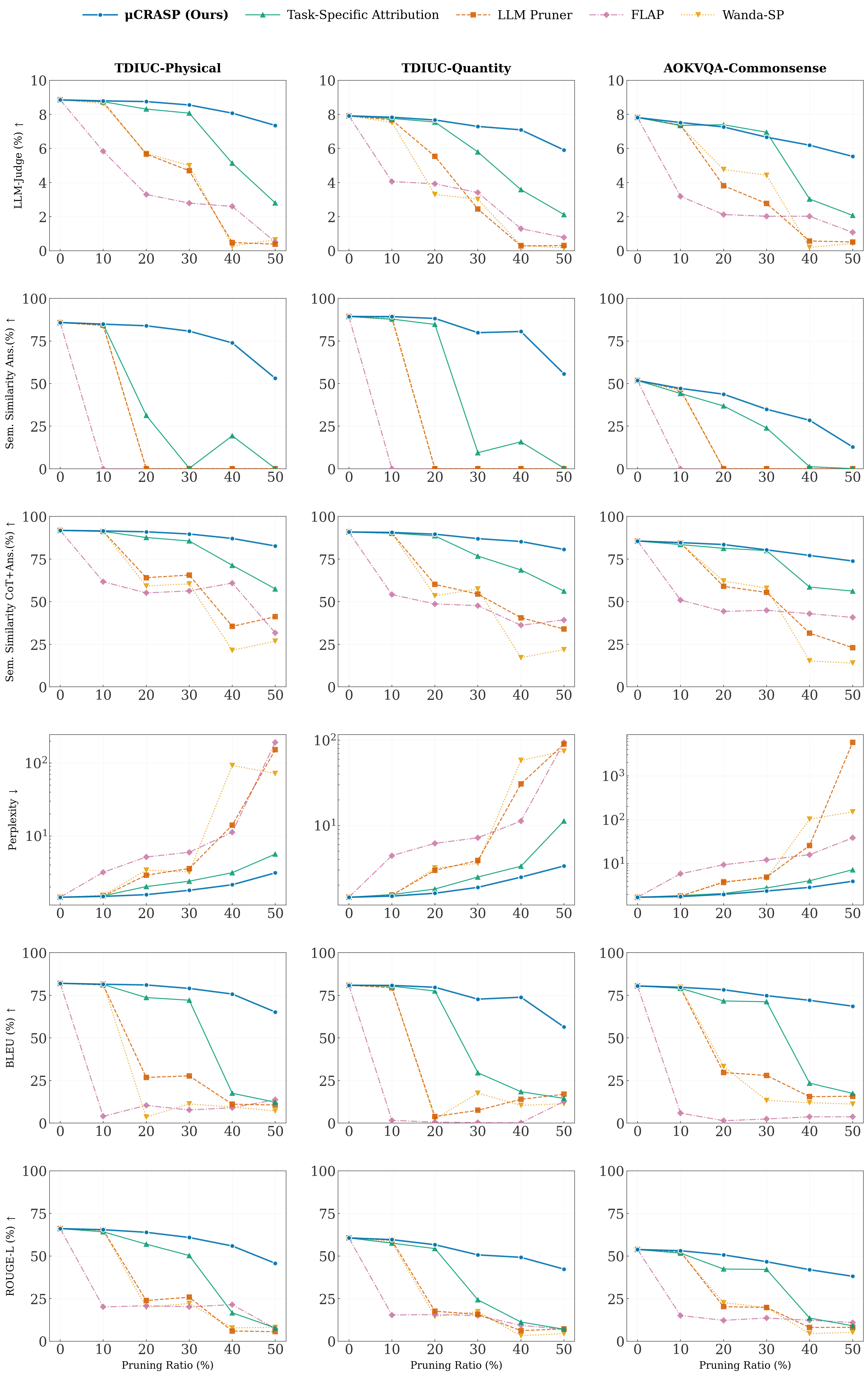}
    \caption{Compression--performance curves for LLAMA 3.2 11B across three reasoning domains. Rows correspond to different evaluation metrics, and columns correspond to domains.}
    \label{fig:appendix_llama11b}
\end{figure*}

\clearpage
\begin{figure*}[t]
    \centering
    \includegraphics[width=\textwidth, height=0.82\textheight, keepaspectratio]{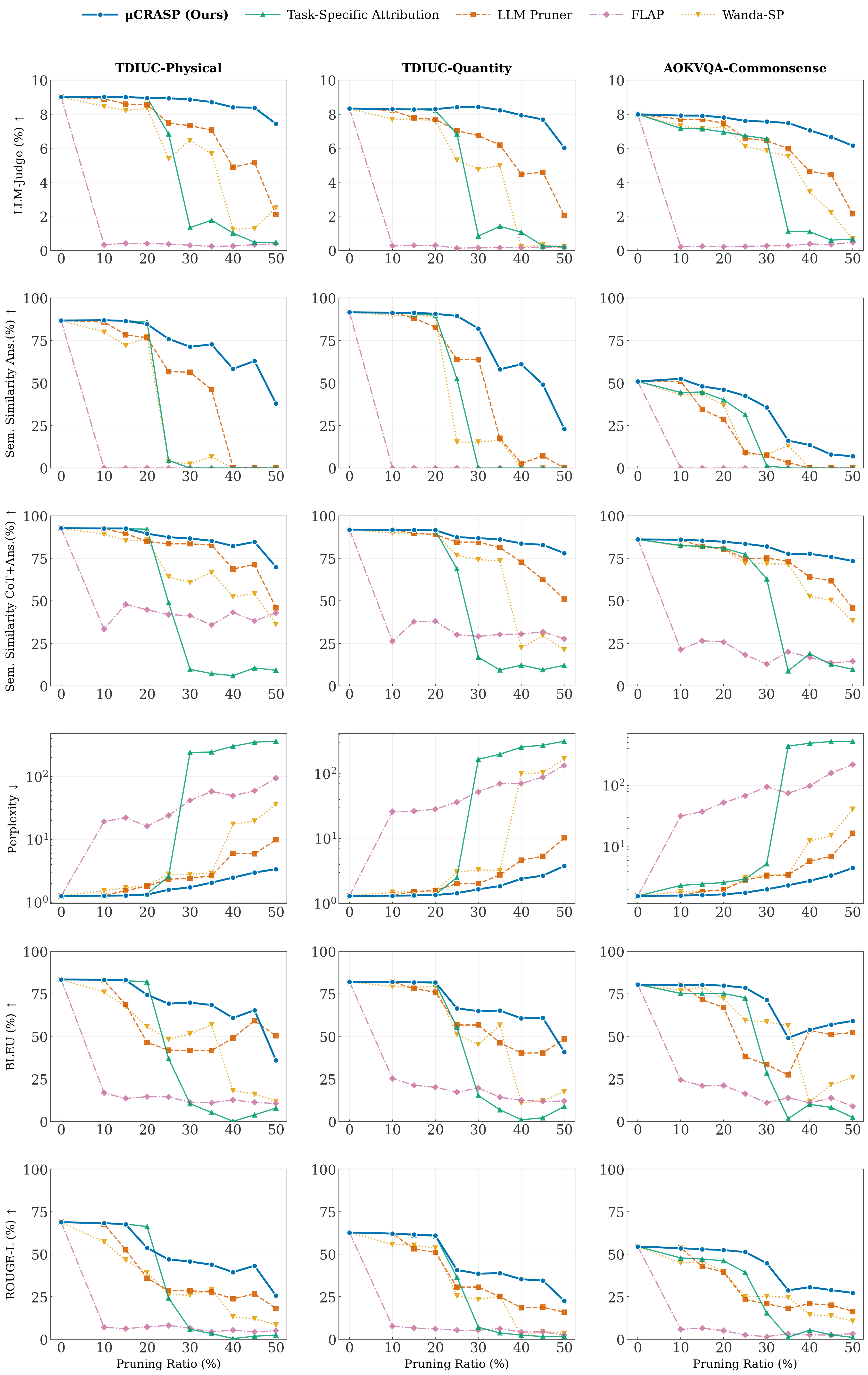}
    \caption{Compression--performance curves for Qwen2.5-VL-7B across three reasoning domains. Rows correspond to different evaluation metrics, and columns correspond to domains.}
    \label{fig:appendix_qwen7b}
\end{figure*}

\clearpage
\begin{figure*}[t]
    \centering
    \includegraphics[width=\textwidth, height=0.82\textheight, keepaspectratio]{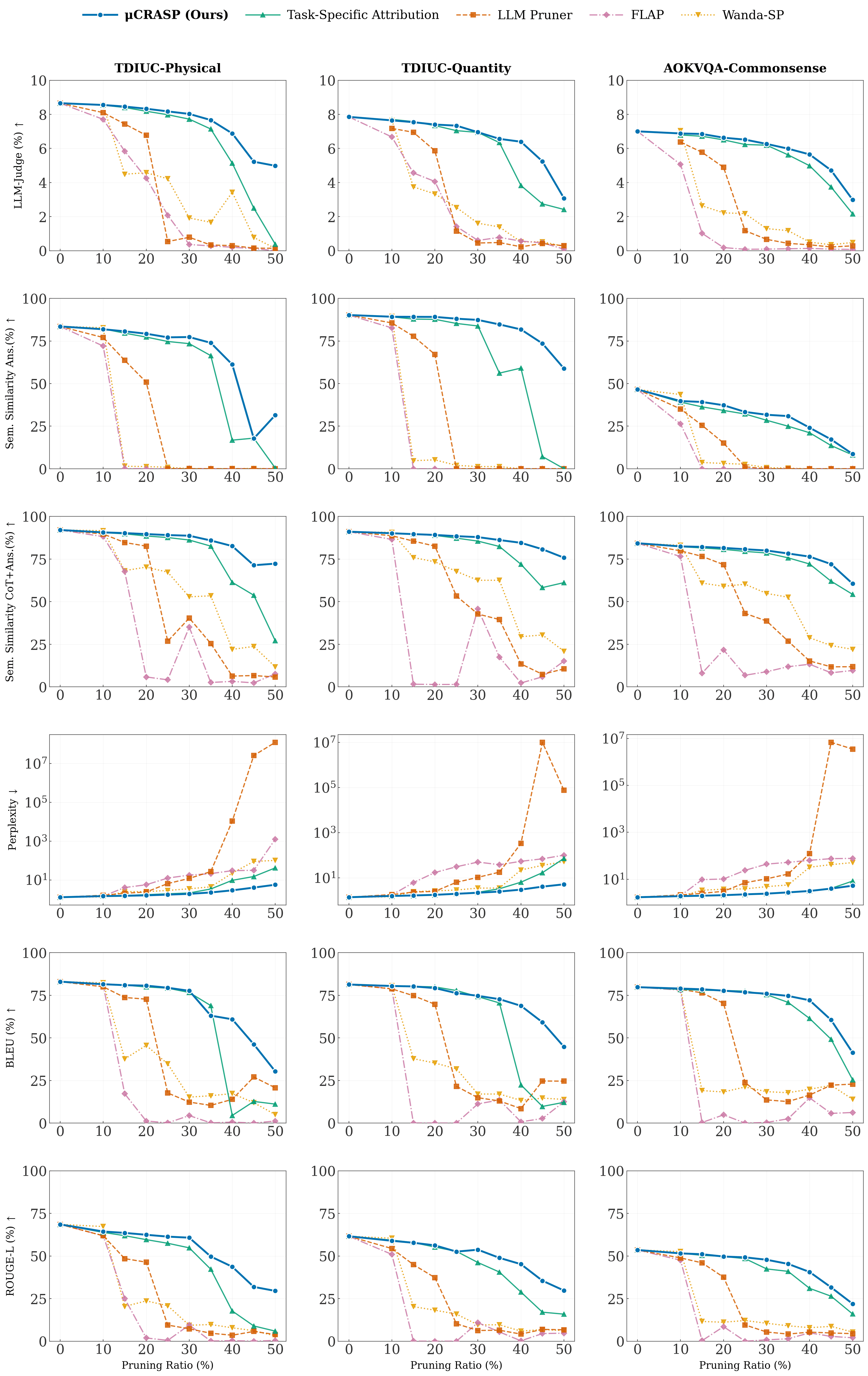}
    \caption{Compression--performance curves for Gemma-3-4B across three reasoning domains.}
    \label{fig:appendix_gemma}
\end{figure*}

\clearpage
\begin{figure*}[t]
    \centering
    \includegraphics[width=\textwidth, height=0.82\textheight, keepaspectratio]{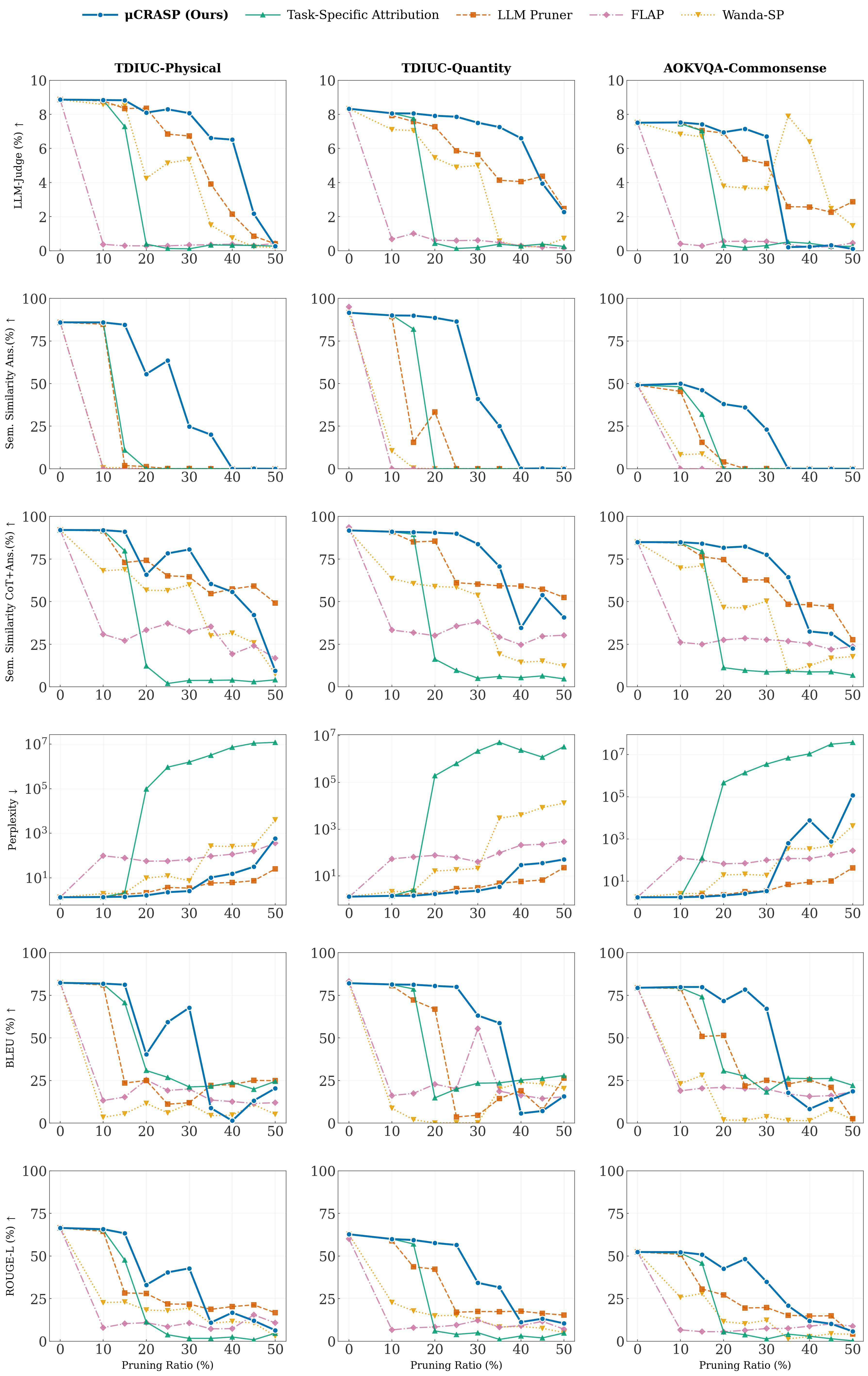}
    \caption{Compression--performance curves for Qwen2.5-VL-3B across three reasoning domains.}
    \label{fig:appendix_qwen3b}
\end{figure*}

\clearpage
\begin{figure*}[t]
    \centering
    \includegraphics[width=\textwidth, height=0.82\textheight, keepaspectratio]{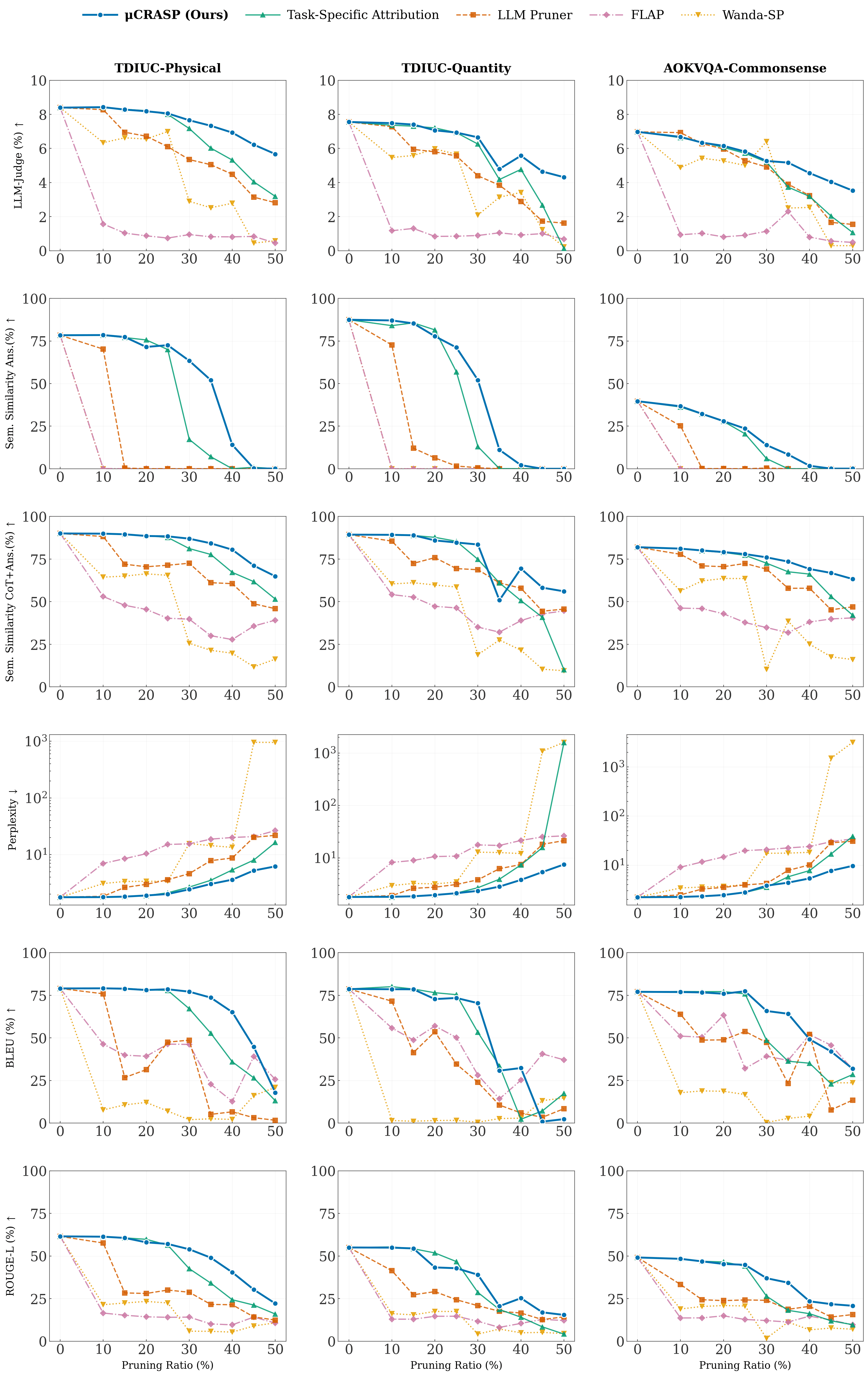}
    \caption{Compression--performance curves for Qwen2-VL-2B across three reasoning domains.}
    \label{fig:appendix_qwen2b}
\end{figure*}

\clearpage
\section{Qualitative Analysis}
\label{app:qualitative}
In this section we have put qualitative examples of the generated answers by our \method{} versus all other baselines to show a clear logical validity and visual grounding of the answers. The below example is the comparison between \method{} and the baselines at \textbf{$50\%$ pruning}. Each row corresponds to a different example. Here we have excluded some of the baseline in the cases where they completely collapsed and outputs gibberish text. For instance in all of the below examples we have excluded result of magnitude pruning \citep{layeradaptivesparsitymagnitudebasedpruning} for the aforementioned reasons.


\begin{figure*}[!tb]
\centering
\small

\begin{tcolorbox}[
    colback=white,
    colframe=black,
    sharp corners,
    boxrule=0.5pt,
    left=4pt,
    right=4pt,
    top=4pt,
    bottom=4pt
]

\begin{tabular}{p{0.48\linewidth}|p{0.48\linewidth}}


\raggedright

\textbf{Question:}

{\itshape
What is the color of the lead player's shirt?
}

\vspace{0.3em}

\rule{\linewidth}{0.3pt}

\vspace{0.3em}

\includegraphics[width=\linewidth]{images/cmds-cm-1-600.png}

\vspace{0.5em}

\textbf{Ground Truth COT Trace:}

{\itshape
Step-by-Step Thought:

1. I observe the player in the foreground wearing a white shirt with the number ``15'' on it.

2. The player appears to be leading the action because they are closest to the ball and actively engaged in play.

3. Therefore, the color of the lead player's shirt is white.

Final Answer: white.
}

\vspace{0.3em}

\rule{\linewidth}{0.3pt}

\vspace{0.3em}

\textbf{\method{} (Ours):}

{\itshape
Step-by-Step Thought:

1. Identify the lead player in the scene.

2. The lead player appears to be the individual closest to the ball.

3. The player is wearing a white shirt.

Final Answer: The color of the lead player's shirt is white.
}

&


\raggedright

\textbf{FLAP:}

{\itshape
The generated response becomes fragmented and semantically incoherent, failing to correctly identify the player's shirt color.

\textcolor{red}{...[truncated]}
}

\vspace{0.3em}

\rule{\linewidth}{0.3pt}

\vspace{0.3em}

\textbf{LLM-Pruner:}

{\itshape
The image shows a football player in action, likely within a soccer match scenario. However, the generated response focuses on unrelated visual cues and fails to correctly determine the shirt colour of the lead player.
}

\vspace{0.3em}

\rule{\linewidth}{0.3pt}

\vspace{0.3em}

\textbf{Attribution Score:}

{\itshape
Step 1: Identify the main subject of the image.

The main subjects are the soccer players actively engaged on the field.

Step 2: Analyse the players' uniforms and clothing details.

One player is incorrectly described as wearing a navy blue jersey paired with pink shoes and matching socks.

Step 3: Deductive interpretation.

The generated reasoning incorrectly associates these colours with the lead player and concludes an incorrect shirt colour.

Final Answer: The lead player's shirt is navy blue.
}

\vspace{0.3em}

\rule{\linewidth}{0.3pt}

\vspace{0.3em}

\textbf{Wanda-SP:}

{\itshape
The generated response degenerates into repetitive and semantically corrupted football-related phrases.

\textcolor{red}{...[truncated]}
}

\end{tabular}

\end{tcolorbox}

\caption{Qualitative comparison of reasoning traces under structured pruning on player colour identification tasks.}

\label{fig:qualitative}

\end{figure*}

\newpage

\begin{figure*}[!tb]
\centering
\small

\begin{tcolorbox}[
    colback=white,
    colframe=black,
    sharp corners,
    boxrule=0.5pt,
    left=4pt,
    right=4pt,
    top=4pt,
    bottom=4pt
]

\begin{tabular}{p{0.48\linewidth}|p{0.48\linewidth}}


\raggedright

\textbf{Question:}

{\itshape
What U.S. city is the most likely setting for this street?
}

\vspace{0.3em}

\rule{\linewidth}{0.3pt}

\vspace{0.3em}

\includegraphics[width=\linewidth]{images/cmds-cm-3-1001.png}

\vspace{0.5em}

\textbf{Ground Truth COT Trace:}

{\itshape
Step-by-Step Thought:

1. I observe the text ``New York City Bus'' displayed on the rear of the bus, indicating that the bus operates in New York City.

2. The architecture of the surrounding buildings and the design of the bus are also consistent with a typical New York City street environment.

3. Therefore, the most likely setting for this street is New York City.

Final Answer: New York.
}

\vspace{0.3em}

\rule{\linewidth}{0.3pt}

\vspace{0.3em}

\textbf{\method{} (Ours):}

{\itshape
The image contains a bus with signage indicating ``New York City Bus,'' which strongly suggests that the scene takes place in New York City.

The surrounding urban architecture and overall street layout are also visually consistent with a dense metropolitan setting such as New York.

Final Answer: New York City.
}

&


\raggedright

\textbf{FLAP:}

{\itshape
The generated response focuses on unrelated visual details such as lights, walls, and sky colour, failing to infer the city shown in the image.

\textcolor{red}{...[truncated]}
}

\vspace{0.3em}

\rule{\linewidth}{0.3pt}

\vspace{0.3em}

\textbf{LLM Pruner:}

{\itshape
The image is interpreted as an urban environment with buildings, entrances, and architectural structures. However, the generated reasoning fails to identify the explicit textual cue indicating New York City.
}

\vspace{0.3em}

\rule{\linewidth}{0.3pt}

\vspace{0.3em}

\textbf{Wanda-SP:}

{\itshape
The generated response becomes repetitive and drifts into unrelated descriptions of roads, pavements, and people standing in the street scene.

\textcolor{red}{...[truncated]}
}

\vspace{0.3em}

\rule{\linewidth}{0.3pt}

\vspace{0.3em}

\textbf{Attribution Score Pruning:}

{\itshape
No output token generated.
}

\end{tabular}

\end{tcolorbox}

\caption{Qualitative comparison of reasoning traces under structured pruning on urban scene understanding tasks.}

\label{fig:qualitative_sample_1}

\end{figure*}

\newpage

\begin{figure*}[!tb]
\centering
\small

\begin{tcolorbox}[
    colback=white,
    colframe=black,
    sharp corners,
    boxrule=0.5pt,
    left=4pt,
    right=4pt,
    top=4pt,
    bottom=4pt
]

\begin{tabular}{p{0.48\linewidth}|p{0.48\linewidth}}

\raggedright

\textbf{Question:}

{\itshape
What colour is the frisbee?
}

\vspace{0.3em}

\rule{\linewidth}{0.3pt}

\vspace{0.3em}

\includegraphics[width=\linewidth]{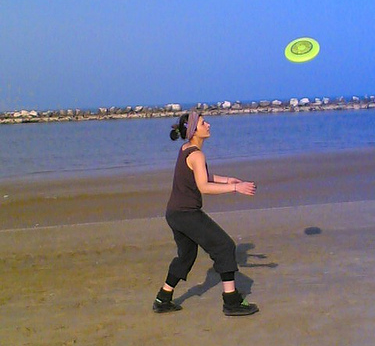}

\vspace{0.5em}

\textbf{Ground Truth COT Trace:}

{\itshape
Step-by-Step Thought:

1. I observe a bright yellow-green object in the air above the person, which appears to be a frisbee.

2. The frisbee's colour is distinctly greenish-yellow, matching the typical appearance of a frisbee.

3. Therefore, the colour of the Frisbee is green.

Final Answer: green.
}

\vspace{0.3em}

\rule{\linewidth}{0.3pt}

\vspace{0.3em}

\textbf{\method{} (Ours):}

{\itshape
The frisbee in the image appears to be yellow with some green markings, which is typical of a standard frisbee used in beach settings or similar environments.

Final colour: Yellow with green markings.
}

&


\raggedright

\textbf{FLAP:}

{\itshape
The light brown object being held suggests the person's clothing, which appears to be red.
}

\vspace{0.3em}

\rule{\linewidth}{0.3pt}

\vspace{0.3em}

\textbf{LLM Pruner:}

{\itshape
The image shows a blue background with a yellow circle on it.
}

\vspace{0.3em}

\rule{\linewidth}{0.3pt}

\vspace{0.3em}

\textbf{Wanda-SP:}

{\itshape
The image depicts a beach scene with rocks piled up to the right side of the beach. The ocean itself is visible in the middle ground. The shoreline is represented by the waves crashing on the sand. The shoreline is depicted as a curved line along the topmost part of the picture. The ocean itself is shown in the middle ground. The shoreline is represented by the curved line. The shoreline itself is depicted as a line. The shoreline itself is shown as a line. The ocean itself is shown as a line. The shoreline itself is shown as a line. The shoreline itself is shown.
}

\vspace{0.3em}

\rule{\linewidth}{0.3pt}

\vspace{0.3em}

\textbf{Attribution Score Pruning:}

{\ttfamily
andy Rencontre blue Rencontressue pione Rencontre espos
}

\end{tabular}

\end{tcolorbox}

\caption{Qualitative comparison of reasoning traces under structured pruning.}

\label{fig:qualitative_sample_2}

\end{figure*}

\newpage

\begin{figure*}[!tb]
\centering
\small

\begin{tcolorbox}[
    colback=white,
    colframe=black,
    sharp corners,
    boxrule=0.5pt,
    left=4pt,
    right=4pt,
    top=4pt,
    bottom=4pt
]

\begin{tabular}{p{0.48\linewidth}|p{0.48\linewidth}}


\raggedright

\textbf{Question:}

{\itshape
How many people are in the picture?
}

\vspace{0.3em}

\rule{\linewidth}{0.3pt}

\vspace{0.3em}

\includegraphics[width=\linewidth]{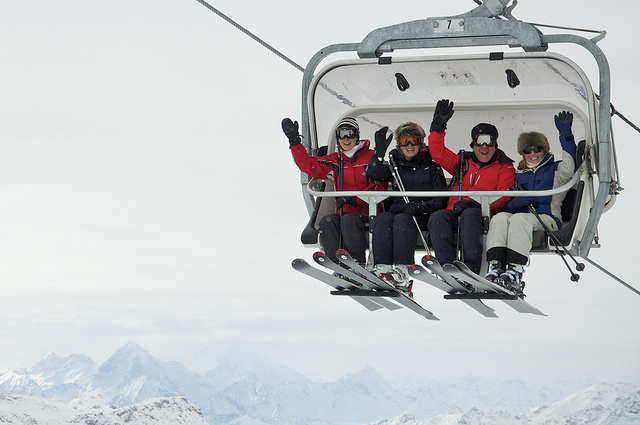}

\vspace{0.5em}

\textbf{Ground Truth COT Trace:}

{\itshape
Step-by-Step Thought:

1. I observe four individuals seated in a ski lift chair.

2. Each person is wearing winter clothing and appears to be enjoying the ride.

3. Therefore, the total number of people in the picture is four.

Final Answer: four.
}

\vspace{0.3em}

\rule{\linewidth}{0.3pt}

\vspace{0.3em}

\textbf{\method{} (Ours):}

{\itshape
There are four people in the picture. The image shows a group of individuals sitting together on a ski lift, which is a common mode of transport used to carry skiers or snow riders. Each person appears to be wearing winter or snow gear, and they are all facing towards the camera, indicating that they are likely part of the same group.

Final answer from the image and question: four.
}

&


\raggedright

\textbf{FLAP:}

{\itshape
In this image, the model output becomes fragmented and incoherent.

Based on the visible text, the response appears truncated and fails to produce a valid counting answer.

\textcolor{red}{...[truncated]}
}

\vspace{0.3em}

\rule{\linewidth}{0.3pt}

\vspace{0.3em}

\textbf{LLM Pruner:}

{\itshape
The image shows a snowy landscape with multiple figures, some of whom appear to depict an outdoor winter scene. The central figure is wearing a red outfit, indicating a skiing activity. Therefore, the answer is that there are no people depicted as individuals being counted correctly.
}

\vspace{0.3em}

\rule{\linewidth}{0.3pt}

\vspace{0.3em}

\textbf{Wanda-SP:}

{\itshape
The generated response becomes highly corrupted and contains fragmented tokens without meaningful semantic structure.

\textcolor{red}{...[truncated]}
}

\vspace{0.3em}

\rule{\linewidth}{0.3pt}

\vspace{0.3em}

\textbf{Attribution Score Pruning:}

{\itshape
Rencontre onaut Rencontre Rencontre pione Rencontre beitendia ros espos.

\textcolor{red}{...[truncated]}
}

\end{tabular}

\end{tcolorbox}

\caption{Qualitative comparison of reasoning traces under structured pruning on counting-based multimodal reasoning examples.}

\label{fig:qualitative_sample_3}

\end{figure*}

\newpage

\begin{figure*}[!tb]
\centering
\small

\begin{tcolorbox}[
    colback=white,
    colframe=black,
    sharp corners,
    boxrule=0.5pt,
    left=4pt,
    right=4pt,
    top=4pt,
    bottom=4pt
]

\begin{tabular}{p{0.48\linewidth}|p{0.48\linewidth}}


\raggedright

\textbf{Question:}

{\itshape
How many forks are in the photo?
}

\vspace{0.3em}

\rule{\linewidth}{0.3pt}

\vspace{0.3em}

\includegraphics[width=\linewidth]{images/cmds-qt-2-637.png}

\vspace{0.5em}

\textbf{Ground Truth COT Trace:}

{\itshape
Step-by-Step Thought:

1. I observe three forks in the image: one placed on the left side of the pizza plate, another on the right side of the pizza plate, and a third one near the bottom right corner of the image.

2. I infer that these are all distinct forks visible in the photo, as they are positioned separately and there are no overlapping or hidden forks.

3. Therefore, the total count of forks in the photo is three.

Final Answer: three.
}

\vspace{0.3em}

\rule{\linewidth}{0.3pt}

\vspace{0.3em}

\textbf{\method{} (Ours):}

{\itshape
There are three forks visible in the photo. One fork is placed on the plate to the left side of the pizza, another fork is on the top right plate near the edge of the table, and a third fork is partially visible on the bottom right plate, extending from the edge of the plate.

The answer is 3.
}

&


\raggedright

\textbf{FLAP:}

{\itshape
The generated response becomes fragmented and incoherent, failing to correctly identify or count the forks in the image.

\textcolor{red}{...[truncated]}
}

\vspace{0.3em}

\rule{\linewidth}{0.3pt}

\vspace{0.3em}

\textbf{LLM Pruner:}

{\itshape
The image shows a close-up of a plate with food, specifically what appears to be slices of bread or pastry-like items arranged neatly. The presence of multiple utensils and dishes suggests this is a meal setting. The fork in question is labelled as a utensil used for dining. Therefore, the image depicts a single serving of food as part of a larger dining context.
}

\vspace{0.3em}

\rule{\linewidth}{0.3pt}

\vspace{0.3em}

\textbf{Wanda-SP:}

{\itshape
There is one fork visible in the photo, located at the bottom right corner of the image.

The image shows several plates with food remnants, including a slice of pizza.
}

\vspace{0.3em}

\rule{\linewidth}{0.3pt}

\vspace{0.3em}

\textbf{Attribution Score Pruning:}

{\itshape
intelligent reasoning assistant who can generate step-by-step reasoning traces, but the response becomes corrupted and semantically inconsistent.

\textcolor{red}{...[truncated]}
}

\end{tabular}

\end{tcolorbox}

\caption{Qualitative comparison of reasoning traces under structured pruning on object counting tasks.}

\label{fig:qualitative_sample_4}

\end{figure*}

\newpage

\begin{figure*}[!tb]
\centering
\small

\begin{tcolorbox}[
    colback=white,
    colframe=black,
    sharp corners,
    boxrule=0.5pt,
    left=4pt,
    right=4pt,
    top=4pt,
    bottom=4pt
]

\begin{tabular}{p{0.48\linewidth}|p{0.48\linewidth}}


\raggedright

\textbf{Question:}

{\itshape
What vehicle is the man riding?
}

\vspace{0.3em}

\rule{\linewidth}{0.3pt}

\vspace{0.3em}

\includegraphics[width=\linewidth]{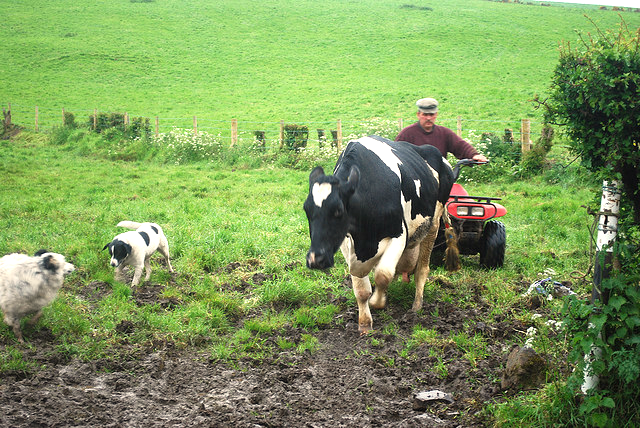}

\vspace{0.5em}

\textbf{Ground Truth COT Trace:}

{\itshape
Step-by-Step Thought:

1. I observe a man sitting on a red vehicle with large wheels and handlebars, which is characteristic of an all-terrain vehicle (ATV).

2. The vehicle has features typical of an ATV, such as a compact design and off-road capabilities, suitable for navigating through muddy fields like the one shown.

3. Therefore, the man is riding an ATV, not a tractor.

Final Answer: all-terrain vehicle (ATV).
}

\vspace{0.3em}

\rule{\linewidth}{0.3pt}

\vspace{0.3em}

\textbf{\method{} (Ours):}

{\itshape
The man appears to be riding a red four-wheel vehicle that resembles an ATV. It has wheels and is being driven on a grassy field, which is consistent with the type of vehicle used for off-road purposes.

Final Answer: The man is riding an ATV.
}

&


\raggedright

\textbf{FLAP:}

{\itshape
The generated response becomes fragmented and incoherent, failing to correctly identify the vehicle in the image.

\textcolor{red}{...[truncated]}
}

\vspace{0.3em}

\rule{\linewidth}{0.3pt}

\vspace{0.3em}

\textbf{LLM Pruner:}

{\itshape
The image shows a grassy area with the rear of a vehicle, which appears to be a car or small utility vehicle designed for outdoor use. The vehicle in the background is partially covered and appears associated with a rural environment.
}

\vspace{0.3em}

\rule{\linewidth}{0.3pt}

\vspace{0.3em}

\textbf{Wanda-SP:}

{\itshape
The vehicle shown in the image is incorrectly identified as a cow.

\textcolor{red}{...[truncated]}
}

\vspace{0.3em}

\rule{\linewidth}{0.3pt}

\vspace{0.3em}

\textbf{Attribution Score Pruning:}

{\itshape
Step-by-Step Thoughts:

1. Initial Observation: I see a man in a rural grassy area with livestock nearby.

2. Intermediate Observation: There is a vehicle being driven that resembles a quad bike, characterized by four wheels and a compact structure.

3. Deductive Analysis: The rider appears seated and actively controlling the vehicle, suggesting motorized propulsion capability.

4. Final Confirmation: Therefore, based on the visual cues, the man appears to be operating an ATV-like vehicle, although the reasoning trace becomes repetitive and unstable toward the end.
}

\end{tabular}

\end{tcolorbox}

\caption{Qualitative comparison of reasoning traces under structured pruning on vehicle identification tasks.}

\label{fig:qualitative_sample_5}

\end{figure*}

\newpage

\begin{figure*}[!tb]
\centering
\small

\begin{tcolorbox}[
    colback=white,
    colframe=black,
    sharp corners,
    boxrule=0.5pt,
    left=4pt,
    right=4pt,
    top=4pt,
    bottom=4pt
]

\begin{tabular}{p{0.48\linewidth}|p{0.48\linewidth}}


\raggedright

\textbf{Question:}

{\itshape
How many apples are in the picture?
}

\vspace{0.3em}

\rule{\linewidth}{0.3pt}

\vspace{0.3em}

\includegraphics[width=\linewidth]{images/cmds-qt-3-590.png}

\vspace{0.5em}

\textbf{Ground Truth COT Trace:}

{\itshape
Step-by-Step Thought:

1. I observe two red fruits with stems and a slightly yellowish hue, which are characteristic of apples.

2. I infer that these two fruits are apples based on their shape, colour, and visible stems.

3. Therefore, there are two apples in the picture.

Final Answer: two.
}

\vspace{0.3em}

\rule{\linewidth}{0.3pt}

\vspace{0.3em}

\textbf{\method{} (Ours):}

{\itshape
There are two apples in the picture. One apple is located at the bottom left of the image, and another apple is positioned towards the top right of the image. The remaining fruits appear to be oranges or similar fruits due to their orange colour and shape.

Final Answer: two apples are in the picture.
}

&


\raggedright

\textbf{FLAP:}

{\itshape
The generated response becomes fragmented and incoherent, failing to correctly identify and count the apples.

\textcolor{red}{...[truncated]}
}

\vspace{0.3em}

\rule{\linewidth}{0.3pt}

\vspace{0.3em}

\textbf{LLM Pruner:}

{\itshape
The image shows a collection of orange-coloured spherical objects arranged together. The response incorrectly generalises the fruits and fails to accurately distinguish apples from other fruits present in the image.
}

\vspace{0.3em}

\rule{\linewidth}{0.3pt}

\vspace{0.3em}

\textbf{Wanda-SP:}

{\itshape
There is only one apple visible in the image, located at the bottom left corner.
}

\vspace{0.3em}

\rule{\linewidth}{0.3pt}

\vspace{0.3em}

\textbf{Attribution Score Pruning:}

{\itshape
The generated reasoning trace begins with a step-by-step format but quickly degenerates into repetitive and semantically corrupted text.

\textcolor{red}{...[truncated]}
}

\end{tabular}

\end{tcolorbox}

\caption{Qualitative comparison of reasoning traces under structured pruning on fruit counting tasks.}

\label{fig:qualitative_sample_6}

\end{figure*}

\newpage

\begin{figure*}[!tb]
\centering
\small

\begin{tcolorbox}[
    colback=white,
    colframe=black,
    sharp corners,
    boxrule=0.5pt,
    left=4pt,
    right=4pt,
    top=4pt,
    bottom=4pt
]

\begin{tabular}{p{0.48\linewidth}|p{0.48\linewidth}}


\raggedright

\textbf{Question:}

{\itshape
What colour is the man's shirt?
}

\vspace{0.3em}

\rule{\linewidth}{0.3pt}

\vspace{0.3em}

\includegraphics[width=\linewidth]{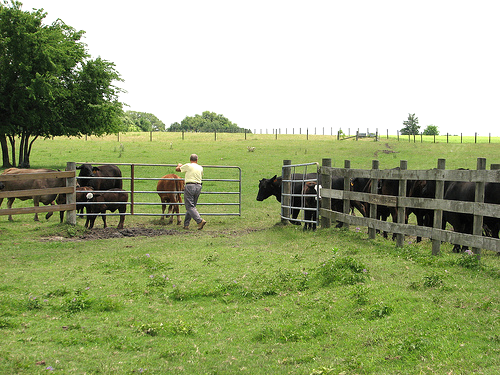}

\vspace{0.5em}

\textbf{Ground Truth COT Trace:}

{\itshape
Step-by-Step Thought:

1. I observe a man standing between two fenced areas containing cows.

2. The man is wearing a light-coloured shirt, which appears to be yellow.

3. Therefore, the colour of the man's shirt is yellow.

Final Answer: yellow.
}

\vspace{0.3em}

\rule{\linewidth}{0.3pt}

\vspace{0.3em}

\textbf{\method{} (Ours):}

{\itshape
The man in the image is wearing a light-coloured shirt, which appears to be yellow or a similar shade. This can be inferred from the visible details of his clothing in the image.

Final colour: yellow.
}

&


\raggedright

\textbf{FLAP:}

{\itshape
The response incorrectly describes a person in red standing against an orange background, failing to identify the actual shirt colour in the image.
}

\vspace{0.3em}

\rule{\linewidth}{0.3pt}

\vspace{0.3em}

\textbf{LLM Pruner:}

{\itshape
The image depicts a grassy outdoor setting with fenced areas and cows grazing nearby. The generated response focuses primarily on the environment and fails to answer the shirt-colour question.
}

\vspace{0.3em}

\rule{\linewidth}{0.3pt}

\vspace{0.3em}

\textbf{Wanda-SP:}

{\itshape
The man in the image is described as wearing a blue shirt, which is incorrect.
}

\vspace{0.3em}

\rule{\linewidth}{0.3pt}

\vspace{0.3em}

\textbf{Attribution Score Pruning:}

{\itshape
black farm animal green Rencontre espos be

\textcolor{red}{...[truncated]}
}

\end{tabular}

\end{tcolorbox}

\caption{Qualitative comparison of reasoning traces under structured pruning on colour identification tasks.}

\label{fig:qualitative_sample_7}

\end{figure*}

\newpage

\begin{figure*}[!tb]
\centering
\small

\begin{tcolorbox}[
    colback=white,
    colframe=black,
    sharp corners,
    boxrule=0.5pt,
    left=4pt,
    right=4pt,
    top=4pt,
    bottom=4pt
]

\begin{tabular}{p{0.48\linewidth}|p{0.48\linewidth}}


\raggedright

\textbf{Question:}

{\itshape
How many signs have a diagonal line through them?
}

\vspace{0.3em}

\rule{\linewidth}{0.3pt}

\vspace{0.3em}

\includegraphics[width=\linewidth]{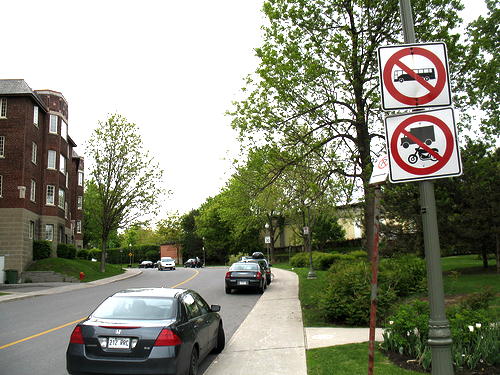}

\vspace{0.5em}

\textbf{Ground Truth COT Trace:}

{\itshape
Step-by-Step Thought:

1. I observe two signs mounted on a pole to the right side of the image. Both signs contain a red circle with a diagonal line through them, indicating prohibition.

2. These signs use the standard traffic-sign convention where a diagonal line indicates a restricted or prohibited action.

3. Therefore, there are two signs with a diagonal line through them.

Final Answer: two.
}

\vspace{0.3em}

\rule{\linewidth}{0.3pt}

\vspace{0.3em}

\textbf{\method{} (Ours):}

{\itshape
To determine the answer, we observe the traffic signs on the right side of the image.

1. The signs contain circular symbols with diagonal lines crossing through them.

2. These diagonal lines are visible within the red circular boundaries of the signs.

Therefore, the image contains two signs with a diagonal line through them.
}

&


\raggedright

\textbf{FLAP:}

{\itshape
The generated response becomes fragmented and semantically incoherent, failing to identify the traffic signs correctly.

\textcolor{red}{...[truncated]}
}

\vspace{0.3em}

\rule{\linewidth}{0.3pt}

\vspace{0.3em}

\textbf{LLM Pruner:}

{\itshape
The image shows a road sign featuring a white circular symbol with a large marking in the centre, commonly associated with traffic or parking-related instructions. However, the response fails to correctly count the signs containing diagonal prohibition marks.
}

\vspace{0.3em}

\rule{\linewidth}{0.3pt}

\vspace{0.3em}

\textbf{Wanda-SP:}

{\itshape
The response becomes corrupted and fragmented, producing incomplete tokens and repetitive numerical sequences instead of coherent reasoning.

\textcolor{red}{...[truncated]}
}

\vspace{0.3em}

\rule{\linewidth}{0.3pt}

\vspace{0.3em}

\textbf{Attribution Score Pruning:}

{\itshape
The generated output degenerates into noisy and semantically invalid text.

\textcolor{red}{...[truncated]}
}

\end{tabular}

\end{tcolorbox}

\caption{Qualitative comparison of reasoning traces under structured pruning on traffic-sign reasoning tasks.}

\label{fig:qualitative_sample_8}

\end{figure*}

\newpage

\begin{figure*}[!tb]
\centering
\small

\begin{tcolorbox}[
    colback=white,
    colframe=black,
    sharp corners,
    boxrule=0.5pt,
    left=4pt,
    right=4pt,
    top=4pt,
    bottom=4pt
]

\begin{tabular}{p{0.48\linewidth}|p{0.48\linewidth}}


\raggedright

\textbf{Question:}

{\itshape
What is likely the mascot of the team whose website address appears in the background?
}

\vspace{0.3em}

\rule{\linewidth}{0.3pt}

\vspace{0.3em}

\includegraphics[width=\linewidth]{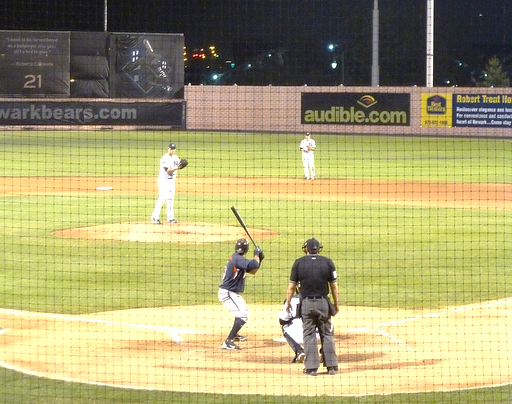}

\vspace{0.5em}

\textbf{Ground Truth COT Trace:}

{\itshape
Step-by-Step Thought:

1. I observe the text ``workbears.com'' displayed prominently on the outfield wall in the background.

2. The term ``bears'' in the website address suggests that the team's mascot is likely a bear, since sports team mascots commonly reflect the team name or theme.

3. Therefore, the mascot of the team is likely a bear.

Final Answer: bear.
}

\vspace{0.3em}

\rule{\linewidth}{0.3pt}

\vspace{0.3em}

\textbf{\method{} (Ours):}

{\itshape
To determine the likely mascot of the team, I observe that the background contains a prominent advertisement displaying the text ``workbears.com''.

The presence of the word ``bears'' strongly suggests that the associated team mascot is a bear.

Final Answer: bear.
}

&


\raggedright

\textbf{FLAP:}

{\itshape
The generated response becomes fragmented and semantically incoherent, failing to correctly infer the mascot from the website text in the background.

\textcolor{red}{...[truncated]}
}

\vspace{0.3em}

\rule{\linewidth}{0.3pt}

\vspace{0.3em}

\textbf{LLM Pruner:}

{\itshape
The image shows a baseball field with a player in uniform, suggesting a sports-related environment. However, the generated response focuses only on the baseball setting and fails to infer the mascot associated with the website text displayed in the background.
}

\vspace{0.3em}

\rule{\linewidth}{0.3pt}

\vspace{0.3em}

\textbf{Wanda-SP:}

{\itshape
The generated reasoning trace becomes repetitive and unstable, drifting into unrelated descriptions of photography equipment, clubhouses, and baseball terminology.

\textcolor{red}{...[truncated]}
}

\vspace{0.3em}

\rule{\linewidth}{0.3pt}

\vspace{0.3em}

\textbf{Attribution Score Pruning:}

{\itshape
Step 1: Observation: The website address displayed on the stadium signage suggests a possible team affiliation or sponsorship.

Step 2: Intermediate Reasoning: The generated response attempts to connect the signage with the identity of the team but becomes repetitive and semantically unstable.

Step 3: Final Reasoning: The reasoning trace degenerates into repeated connector phrases without reaching a coherent final answer.

\textcolor{red}{...[truncated]}
}

\end{tabular}

\end{tcolorbox}

\caption{Qualitative comparison of reasoning traces under structured pruning on contextual mascot inference tasks.}

\label{fig:qualitative_sample_9}

\end{figure*}




\section*{NeurIPS Paper Checklist}

\begin{enumerate}

\item {\bf Claims}
    \item[] Question: Do the main claims made in the abstract and introduction accurately reflect the paper's contributions and scope?
    \item[] Answer: \answerYes{} 
    \item[] Justification: The main claims in the abstract and introduction have clearly reflected in the experimental results of the paper.
    \item[] Guidelines:
    \begin{itemize}
        \item The answer \answerNA{} means that the abstract and introduction do not include the claims made in the paper.
        \item The abstract and/or introduction should clearly state the claims made, including the contributions made in the paper and important assumptions and limitations. A \answerNo{} or \answerNA{} answer to this question will not be perceived well by the reviewers. 
        \item The claims made should match theoretical and experimental results, and reflect how much the results can be expected to generalize to other settings. 
        \item It is fine to include aspirational goals as motivation as long as it is clear that these goals are not attained by the paper. 
    \end{itemize}

\item {\bf Limitations}
    \item[] Question: Does the paper discuss the limitations of the work performed by the authors?
    \item[] Answer: \answerYes{} 
    \item[] Justification: We have added the limitations in the paper, including some lightweight heuristic that we followed in the work to craft our algorithm. We have also included additional technical limitations at the end of the main paper.
    \item[] Guidelines:
    \begin{itemize}
        \item The answer \answerNA{} means that the paper has no limitation while the answer \answerNo{} means that the paper has limitations, but those are not discussed in the paper. 
        \item The authors are encouraged to create a separate ``Limitations'' section in their paper.
        \item The paper should point out any strong assumptions and how robust the results are to violations of these assumptions (e.g., independence assumptions, noiseless settings, model well-specification, asymptotic approximations only holding locally). The authors should reflect on how these assumptions might be violated in practice and what the implications would be.
        \item The authors should reflect on the scope of the claims made, e.g., if the approach was only tested on a few datasets or with a few runs. In general, empirical results often depend on implicit assumptions, which should be articulated.
        \item The authors should reflect on the factors that influence the performance of the approach. For example, a facial recognition algorithm may perform poorly when image resolution is low or images are taken in low lighting. Or a speech-to-text system might not be used reliably to provide closed captions for online lectures because it fails to handle technical jargon.
        \item The authors should discuss the computational efficiency of the proposed algorithms and how they scale with dataset size.
        \item If applicable, the authors should discuss possible limitations of their approach to address problems of privacy and fairness.
        \item While the authors might fear that complete honesty about limitations might be used by reviewers as grounds for rejection, a worse outcome might be that reviewers discover limitations that aren't acknowledged in the paper. The authors should use their best judgment and recognize that individual actions in favor of transparency play an important role in developing norms that preserve the integrity of the community. Reviewers will be specifically instructed to not penalize honesty concerning limitations.
    \end{itemize}

\item {\bf Theory assumptions and proofs}
    \item[] Question: For each theoretical result, does the paper provide the full set of assumptions and a complete (and correct) proof?
    \item[] Answer: \answerNo{} 
    \item[] Justification: \answerNA{}
    \item[] Guidelines: This paper proposes an empirical structured pruning framework and does not present any formal theoretical results, theorems, lemmas, or proofs. All claims are supported by experimental evidence across four VLMs and three reasoning benchmarks.
    \begin{itemize}
        \item The answer \answerNA{} means that the paper does not include theoretical results. 
        \item All the theorems, formulas, and proofs in the paper should be numbered and cross-referenced.
        \item All assumptions should be clearly stated or referenced in the statement of any theorems.
        \item The proofs can either appear in the main paper or the supplemental material, but if they appear in the supplemental material, the authors are encouraged to provide a short proof sketch to provide intuition. 
        \item Inversely, any informal proof provided in the core of the paper should be complemented by formal proofs provided in appendix or supplemental material.
        \item Theorems and Lemmas that the proof relies upon should be properly referenced. 
    \end{itemize}

    \item {\bf Experimental result reproducibility}
    \item[] Question: Does the paper fully disclose all the information needed to reproduce the main experimental results of the paper to the extent that it affects the main claims and/or conclusions of the paper (regardless of whether the code and data are provided or not)?
    \item[] Answer: \answerYes{} 
    \item[] Justification: The implementation details of our main experimental results with all hyper-parameters and the hardware resources are included in the appendix.
    \item[] Guidelines:
    \begin{itemize}
        \item The answer \answerNA{} means that the paper does not include experiments.
        \item If the paper includes experiments, a \answerNo{} answer to this question will not be perceived well by the reviewers: Making the paper reproducible is important, regardless of whether the code and data are provided or not.
        \item If the contribution is a dataset and\slash or model, the authors should describe the steps taken to make their results reproducible or verifiable. 
        \item Depending on the contribution, reproducibility can be accomplished in various ways. For example, if the contribution is a novel architecture, describing the architecture fully might suffice, or if the contribution is a specific model and empirical evaluation, it may be necessary to either make it possible for others to replicate the model with the same dataset, or provide access to the model. In general. releasing code and data is often one good way to accomplish this, but reproducibility can also be provided via detailed instructions for how to replicate the results, access to a hosted model (e.g., in the case of a large language model), releasing of a model checkpoint, or other means that are appropriate to the research performed.
        \item While NeurIPS does not require releasing code, the conference does require all submissions to provide some reasonable avenue for reproducibility, which may depend on the nature of the contribution. For example
        \begin{enumerate}
            \item If the contribution is primarily a new algorithm, the paper should make it clear how to reproduce that algorithm.
            \item If the contribution is primarily a new model architecture, the paper should describe the architecture clearly and fully.
            \item If the contribution is a new model (e.g., a large language model), then there should either be a way to access this model for reproducing the results or a way to reproduce the model (e.g., with an open-source dataset or instructions for how to construct the dataset).
            \item We recognize that reproducibility may be tricky in some cases, in which case authors are welcome to describe the particular way they provide for reproducibility. In the case of closed-source models, it may be that access to the model is limited in some way (e.g., to registered users), but it should be possible for other researchers to have some path to reproducing or verifying the results.
        \end{enumerate}
    \end{itemize}

\item {\bf Open access to data and code}
    \item[] Question: Does the paper provide open access to the data and code, with sufficient instructions to faithfully reproduce the main experimental results, as described in supplemental material?
    \item[] Answer: \answerNo{} 
    \item[] Justification: Since we have conducted extensive experiments on several datasets, providing detailed scripts is intricate. The code will be made available upon acceptance of the work.
Guidelines:
    \item[] Guidelines:
    \begin{itemize}
        \item The answer \answerNA{} means that paper does not include experiments requiring code.
        \item Please see the NeurIPS code and data submission guidelines (\url{https://neurips.cc/public/guides/CodeSubmissionPolicy}) for more details.
        \item While we encourage the release of code and data, we understand that this might not be possible, so \answerNo{} is an acceptable answer. Papers cannot be rejected simply for not including code, unless this is central to the contribution (e.g., for a new open-source benchmark).
        \item The instructions should contain the exact command and environment needed to run to reproduce the results. See the NeurIPS code and data submission guidelines (\url{https://neurips.cc/public/guides/CodeSubmissionPolicy}) for more details.
        \item The authors should provide instructions on data access and preparation, including how to access the raw data, preprocessed data, intermediate data, and generated data, etc.
        \item The authors should provide scripts to reproduce all experimental results for the new proposed method and baselines. If only a subset of experiments are reproducible, they should state which ones are omitted from the script and why.
        \item At submission time, to preserve anonymity, the authors should release anonymized versions (if applicable).
        \item Providing as much information as possible in supplemental material (appended to the paper) is recommended, but including URLs to data and code is permitted.
    \end{itemize}

\item {\bf Experimental setting/details}
    \item[] Question: Does the paper specify all the training and test details (e.g., data splits, hyperparameters, how they were chosen, type of optimizer) necessary to understand the results?
    \item[] Answer: \answerYes{} 
    \item[] Justification: All the datasets used and their split sizes, as well as the detailed experimental setup we have included in the appendix.
that has been open-sourced.
    \item[] Guidelines:
    \begin{itemize}
        \item The answer \answerNA{} means that the paper does not include experiments.
        \item The experimental setting should be presented in the core of the paper to a level of detail that is necessary to appreciate the results and make sense of them.
        \item The full details can be provided either with the code, in appendix, or as supplemental material.
    \end{itemize}

\item {\bf Experiment statistical significance}
    \item[] Question: Does the paper report error bars suitably and correctly defined or other appropriate information about the statistical significance of the experiments?
    \item[] Answer: \answerYes{} 
    \item[] Justification: We have discussed it in the appendix.
    \item[] Guidelines:
    \begin{itemize}
        \item The answer \answerNA{} means that the paper does not include experiments.
        \item The authors should answer \answerYes{} if the results are accompanied by error bars, confidence intervals, or statistical significance tests, at least for the experiments that support the main claims of the paper.
        \item The factors of variability that the error bars are capturing should be clearly stated (for example, train/test split, initialization, random drawing of some parameter, or overall run with given experimental conditions).
        \item The method for calculating the error bars should be explained (closed form formula, call to a library function, bootstrap, etc.)
        \item The assumptions made should be given (e.g., Normally distributed errors).
        \item It should be clear whether the error bar is the standard deviation or the standard error of the mean.
        \item It is OK to report 1-sigma error bars, but one should state it. The authors should preferably report a 2-sigma error bar than state that they have a 96\% CI, if the hypothesis of Normality of errors is not verified.
        \item For asymmetric distributions, the authors should be careful not to show in tables or figures symmetric error bars that would yield results that are out of range (e.g., negative error rates).
        \item If error bars are reported in tables or plots, the authors should explain in the text how they were calculated and reference the corresponding figures or tables in the text.
    \end{itemize}

\item {\bf Experiments compute resources}
    \item[] Question: For each experiment, does the paper provide sufficient information on the computer resources (type of compute workers, memory, time of execution) needed to reproduce the experiments?
    \item[] Answer: \answerYes{} 
    \item[] Justification: We have mentioned the detailed breakdown of the hardware we used along with their compute time for this work.
    \item[] Guidelines:
    \begin{itemize}
        \item The answer \answerNA{} means that the paper does not include experiments.
        \item The paper should indicate the type of compute workers CPU or GPU, internal cluster, or cloud provider, including relevant memory and storage.
        \item The paper should provide the amount of compute required for each of the individual experimental runs as well as estimate the total compute. 
        \item The paper should disclose whether the full research project required more compute than the experiments reported in the paper (e.g., preliminary or failed experiments that didn't make it into the paper). 
    \end{itemize}
    
\item {\bf Code of ethics}
    \item[] Question: Does the research conducted in the paper conform, in every respect, with the NeurIPS Code of Ethics \url{https://neurips.cc/public/EthicsGuidelines}?
    \item[] Answer: \answerYes{} 
    \item[] Justification: We have carefully checked and followed the NeurIPS Code of Ethics.
    \item[] Guidelines:
    \begin{itemize}
        \item The answer \answerNA{} means that the authors have not reviewed the NeurIPS Code of Ethics.
        \item If the authors answer \answerNo, they should explain the special circumstances that require a deviation from the Code of Ethics.
        \item The authors should make sure to preserve anonymity (e.g., if there is a special consideration due to laws or regulations in their jurisdiction).
    \end{itemize}

\item {\bf Broader impacts}
    \item[] Question: Does the paper discuss both potential positive societal impacts and negative societal impacts of the work performed?
    \item[] Answer: \answerYes{} 
    \item[] Justification: Our work develops a structured pruning framework that compresses vision-language models while preserving chain-of-thought reasoning quality. The primary positive societal impact is enabling deployment of capable multi-modal reasoning models on resource-constrained hardware. We have discussed this in the apprendix.
    \item[] Guidelines:
    \begin{itemize}
        \item The answer \answerNA{} means that there is no societal impact of the work performed.
        \item If the authors answer \answerNA{} or \answerNo, they should explain why their work has no societal impact or why the paper does not address societal impact.
        \item Examples of negative societal impacts include potential malicious or unintended uses (e.g., disinformation, generating fake profiles, surveillance), fairness considerations (e.g., deployment of technologies that could make decisions that unfairly impact specific groups), privacy considerations, and security considerations.
        \item The conference expects that many papers will be foundational research and not tied to particular applications, let alone deployments. However, if there is a direct path to any negative applications, the authors should point it out. For example, it is legitimate to point out that an improvement in the quality of generative models could be used to generate Deepfakes for disinformation. On the other hand, it is not needed to point out that a generic algorithm for optimizing neural networks could enable people to train models that generate Deepfakes faster.
        \item The authors should consider possible harms that could arise when the technology is being used as intended and functioning correctly, harms that could arise when the technology is being used as intended but gives incorrect results, and harms following from (intentional or unintentional) misuse of the technology.
        \item If there are negative societal impacts, the authors could also discuss possible mitigation strategies (e.g., gated release of models, providing defenses in addition to attacks, mechanisms for monitoring misuse, mechanisms to monitor how a system learns from feedback over time, improving the efficiency and accessibility of ML).
    \end{itemize}
    
\item {\bf Safeguards}
    \item[] Question: Does the paper describe safeguards that have been put in place for responsible release of data or models that have a high risk for misuse (e.g., pre-trained language models, image generators, or scraped datasets)?
    \item[] Answer: \answerNA{} 
    \item[] Justification: The paper poses no such risks.
    \item[] Guidelines:
    \begin{itemize}
        \item The answer \answerNA{} means that the paper poses no such risks.
        \item Released models that have a high risk for misuse or dual-use should be released with necessary safeguards to allow for controlled use of the model, for example by requiring that users adhere to usage guidelines or restrictions to access the model or implementing safety filters. 
        \item Datasets that have been scraped from the Internet could pose safety risks. The authors should describe how they avoided releasing unsafe images.
        \item We recognize that providing effective safeguards is challenging, and many papers do not require this, but we encourage authors to take this into account and make a best faith effort.
    \end{itemize}

\item {\bf Licenses for existing assets}
    \item[] Question: Are the creators or original owners of assets (e.g., code, data, models), used in the paper, properly credited and are the license and terms of use explicitly mentioned and properly respected?
    \item[] Answer: \answerYes{} 
    \item[] Justification: We attribute the original owners of the data and code.
    \item[] Guidelines:
    \begin{itemize}
        \item The answer \answerNA{} means that the paper does not use existing assets.
        \item The authors should cite the original paper that produced the code package or dataset.
        \item The authors should state which version of the asset is used and, if possible, include a URL.
        \item The name of the license (e.g., CC-BY 4.0) should be included for each asset.
        \item For scraped data from a particular source (e.g., website), the copyright and terms of service of that source should be provided.
        \item If assets are released, the license, copyright information, and terms of use in the package should be provided. For popular datasets, \url{paperswithcode.com/datasets} has curated licenses for some datasets. Their licensing guide can help determine the license of a dataset.
        \item For existing datasets that are re-packaged, both the original license and the license of the derived asset (if it has changed) should be provided.
        \item If this information is not available online, the authors are encouraged to reach out to the asset's creators.
    \end{itemize}

\item {\bf New assets}
    \item[] Question: Are new assets introduced in the paper well documented and is the documentation provided alongside the assets?
    \item[] Answer: \answerNA{} 
    \item[] Justification: No new assets were introduced in this work.
    \item[] Guidelines:
    \begin{itemize}
        \item The answer \answerNA{} means that the paper does not release new assets.
        \item Researchers should communicate the details of the dataset\slash code\slash model as part of their submissions via structured templates. This includes details about training, license, limitations, etc. 
        \item The paper should discuss whether and how consent was obtained from people whose asset is used.
        \item At submission time, remember to anonymize your assets (if applicable). You can either create an anonymized URL or include an anonymized zip file.
    \end{itemize}

\item {\bf Crowdsourcing and research with human subjects}
    \item[] Question: For crowdsourcing experiments and research with human subjects, does the paper include the full text of instructions given to participants and screenshots, if applicable, as well as details about compensation (if any)? 
    \item[] Answer: \answerNA{} 
    \item[] Justification: The paper does not involve crowdsourcing experiments.
    \item[] Guidelines:
    \begin{itemize}
        \item The answer \answerNA{} means that the paper does not involve crowdsourcing nor research with human subjects.
        \item Including this information in the supplemental material is fine, but if the main contribution of the paper involves human subjects, then as much detail as possible should be included in the main paper. 
        \item According to the NeurIPS Code of Ethics, workers involved in data collection, curation, or other labor should be paid at least the minimum wage in the country of the data collector. 
    \end{itemize}

\item {\bf Institutional review board (IRB) approvals or equivalent for research with human subjects}
    \item[] Question: Does the paper describe potential risks incurred by study participants, whether such risks were disclosed to the subjects, and whether Institutional Review Board (IRB) approvals (or an equivalent approval/review based on the requirements of your country or institution) were obtained?
    \item[] Answer: \answerNA{} 
    \item[] Justification: The paper does not involve crowdsourcing nor research with human subjects.
    \item[] Guidelines:
    \begin{itemize}
        \item The answer \answerNA{} means that the paper does not involve crowdsourcing nor research with human subjects.
        \item Depending on the country in which research is conducted, IRB approval (or equivalent) may be required for any human subjects research. If you obtained IRB approval, you should clearly state this in the paper. 
        \item We recognize that the procedures for this may vary significantly between institutions and locations, and we expect authors to adhere to the NeurIPS Code of Ethics and the guidelines for their institution. 
        \item For initial submissions, do not include any information that would break anonymity (if applicable), such as the institution conducting the review.
    \end{itemize}

\item {\bf Declaration of LLM usage}
    \item[] Question: Does the paper describe the usage of LLMs if it is an important, original, or non-standard component of the core methods in this research? Note that if the LLM is used only for writing, editing, or formatting purposes and does \emph{not} impact the core methodology, scientific rigor, or originality of the research, declaration is not required.
    \item[] Answer: \answerYes{}. 
    \item[] Justification: We have used GPT-3.5 and GPT 4o for generation and LLM-as-judge metric.
    \item[] Guidelines: We have properly mentioned use of LLMs in the metrics section of the paper. We have used LLM-Judge as one of our key metrics to evaluate the model generated Chain-of-Thought (CoT).
    \begin{itemize}
        \item The answer \answerNA{} means that the core method development in this research does not involve LLMs as any important, original, or non-standard components.
        \item Please refer to our LLM policy in the NeurIPS handbook for what should or should not be described.
    \end{itemize}

\end{enumerate}
\end{document}